\documentclass[lettersize,journal]{IEEEtran}
\usepackage{amsmath,amsfonts}
\usepackage{array}
\usepackage[caption=false,font=normalsize,labelfont=sf,textfont=sf]{subfig}
\usepackage{textcomp}
\usepackage{stfloats}
\usepackage{url}
\usepackage{hyperref}
\usepackage{verbatim}
\usepackage{graphicx}
\usepackage{cite}
\usepackage[noend]{algpseudocode}
\usepackage[ruled]{algorithm2e}
\usepackage{xcolor, colortbl}
\definecolor{mygray}{gray}{.9}

\makeatletter 
\g@addto@macro{\@algocf@init}{\SetKwInOut{Parameter}{Parameters}} 
\makeatother

\begin{document}

\title{Density Crop-guided Semi-supervised Object Detection in Aerial Images}

%\author{IEEE Publication Technology,~\IEEEmembership{Staff,~IEEE,}}
\author{\IEEEauthorblockN{
        Akhil Meethal,
        Eric granger,        
        Marco Pedersoli
        }\\
      \IEEEauthorblockA{LIVIA lab, Dept. of Systems Engineering, ETS Montreal, Canada}
  }
        % <-this % stops a space
%\thanks{This paper was produced by the IEEE Publication Technology Group. They are in Piscataway, NJ.}% <-this % stops a space
%\thanks{Manuscript received April 19, 2021; revised August 16, 2021.}}

\markboth{Journal of \LaTeX\ Class Files,~Vol.~14, No.~8, August~2021}%
{Shell \MakeLowercase{\textit{et al.}}: A Sample Article Using IEEEtran.cls for IEEE Journals}


\maketitle

% !TEX root=main.tex

\begin{abstract}
One of the important bottlenecks in training modern object detectors is the need for labeled images where bounding box annotations have to be produced for each object present in the image. This bottleneck is further exacerbated in aerial images where the annotators have to label small objects often distributed in clusters on high-resolution images. In recent days, the mean-teacher approach trained with pseudo-labels and weak-strong augmentation consistency is gaining popularity for semi-supervised object detection. However, a direct adaptation of such semi-supervised detectors for aerial images where small clustered objects are often present, might not lead to optimal results. In this paper, we propose a density crop-guided semi-supervised detector that identifies the cluster of small objects during training and also exploits them to improve performance at inference. During training, image crops of clusters identified from labeled and unlabeled images are used to augment the training set, which in turn increases the chance of detecting small objects and creating good pseudo-labels for small objects on the unlabeled images. During inference, the detector is not only able to detect the objects of interest but also regions with a high density of small objects (density crops) so that detections from the input image and detections from image crops are combined, resulting in an overall more accurate object prediction, especially for small objects. Empirical studies on the popular benchmarks of VisDrone and DOTA datasets show the effectiveness of our density crop-guided semi-supervised detector with an average improvement of more than 2\% over the basic mean-teacher method in COCO style AP. Our code is available at: \href{https://github.com/akhilpm/DroneSSOD}{https://github.com/akhilpm/DroneSSOD}
\end{abstract}

% !TEX root=main.tex

\section{Introduction}
\label{sec:introduction}
With abundant labeled data and efficient deep learning algorithms, supervised object detection has achieved impressive results in natural images \cite{fcos-Tian-2019, detr-Carion-2020, vit-dosovitskiy-2020, swin-Li-2021}. Though the progress is also resonated in Aerial Image object detection with images captured by drones and satellites\cite{xia-DOTA-2018, aerial1-Cheng-2016, aerial2-Han-2021, remote_sense-Long-2017}, we are yet to see large-scale annotated datasets like Pascal VOC \cite{voc-Everingham-2010}, MS-COCO \cite{mscoco-Lin-2014} and Open Images\cite{openimages-Alina-2020} in aerial images. Getting sufficient labeled data is difficult in aerial images, especially at instance-level recognition tasks like object detection \cite{soft-teacher-Xu-2021, unbiased-teacher-Liu-2021, semiwsod-Meethal-2022}, limiting the scalability of the popular supervised detectors to aerial images. The annotation of several tiny objects per image is a tedious task \cite{zhu-VisDrone-2018}. 
%The annotation challenge is humongous here as one has to label hundreds of tiny objects per image \cite{zhu-VisDrone-2018}. 
This puts more demand on learning object detectors with limited annotations in aerial image detection. Practical applications with aerial imagery produce large amounts of unlabeled data \cite{uav_data-Caillouet-2019, uav_road_car-Yueming-2022, sat_data-Song-2021} but they are simply not utilized in the learning process. Therefore, it is important to leverage the available unlabeled data to train the detectors for aerial image detection.

\begin{figure*}[!t]
\captionsetup[subfigure]{labelformat=empty}
\centering
\subfloat[]{\includegraphics[width=0.49\textwidth]{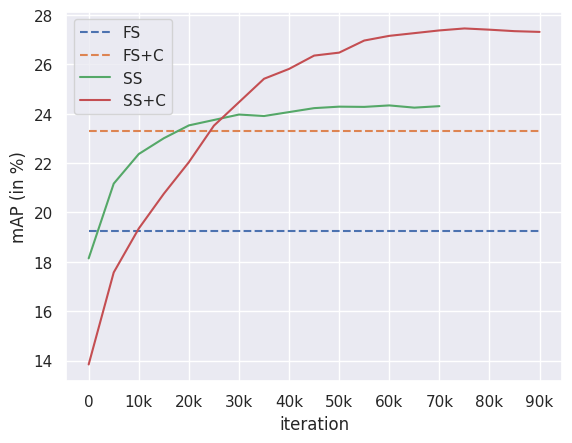}%
\label{fig:ap_over_iter}}
\hfil
\subfloat[]{\includegraphics[width=0.49\textwidth]{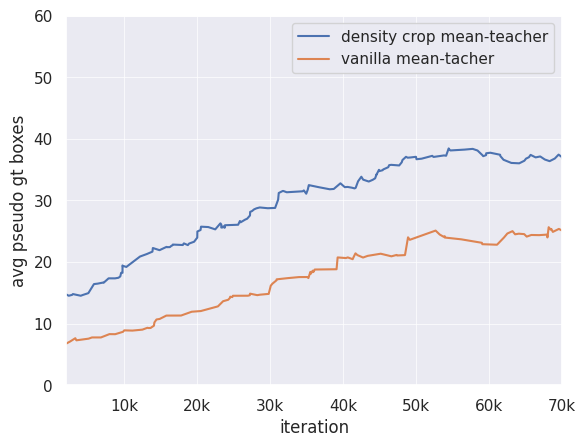}%
\label{number_pseudo_GT_boxes}}
\caption{(a) Change in mAP over the epochs with and without density crops on supervised and semi-supervised settings. FS: Fully Supervised, FS+C: Fully supervised + density crops, SS: Semi-supervised (mean-teacher baseline), SS+C: Semi-supervised + density crops (on labeled and unlabeled images).(b) The average number of pseudo-GT boxes per image over training iteration. The density crop-guided mean-teacher is producing more pseudo labels compared to the vanilla mean-teacher method. This will result in more pseudo-labels for small objects.}
\label{fig:pseudo_gt_ap_over_iter}
\end{figure*}

To train a detector with limited annotated images and a large collection of unlabeled data, we focus on training them in a semi-supervised setting. Although semi-supervised object detection (SSOD) has achieved tremendous progress in recent years on natural images \cite{sed-Guo-2022, class-prototype-Li-2022, unbiased-teacher-Liu-2021, soft-teacher-Xu-2021, humble-teacher-Yang-2021, csd-Jeong-2019, stac-sohn-2015, semiwsod-Meethal-2022}, we are yet to see large-scale adoption of them on aerial images. While adapting the best SSOD techniques on natural images to aerial images, one should be cautious due to the large difference in these images. Specifically, factors like the high-resolution nature of the imagery, the small size of the objects, and their sparse distribution across the image need consideration. The number of target objects is also fairly high in aerial images compared to natural images. For example, the average number of objects in Pascal VOC and MS-COCO images are 3 and 7, respectively, whereas the images in the VisDrone \cite{zhu-VisDrone-2018} and DOTA \cite{xia-DOTA-2018} datasets, two popular benchmarks in aerial detection research, have an average number of 53 and 67 objects, respectively. A naive adaptation of semi-supervised detectors based on pseudo-labels, in this case, is not labeling enough small objects in the unlabeled images as shown in figure \ref{fig:pseudo_gt_ap_over_iter} right. We hypothesize that these difficulties contributed to the lower adoption rate of semi-supervised detectors in aerial image detection. To the best of our knowledge, we are yet to see a large-scale study of the recent semi-supervised detectors in aerial images.

In supervised settings, even with a fully annotated training set, vanilla object detectors still struggle to accurately localize the small objects in high-resolution aerial images. Thus additional techniques for performance enhancement are often used including density-guided detection
\cite{dmap-Li-2020, dmap-Duan-2021, glsan-Deng-2020}, scale-specific detectors \cite{snip-Singh-2018, sniper-Singh-2018}, feature fusion \cite{fpn-Lin-2017, panet-Liu-2018}, attention methods \cite{querydet-Yang-2022} etc. Among these, density crop-based approaches are popular as they process the cluster of small objects by zooming in on the crops obtained from there, resulting in better small object detection. To bring the same benefits to the semi-supervised settings, we designed our semi-supervised detector with density crop-guided training and inference. Although density crops can be used in supervised settings with external learnable modules and additional loss functions \cite{dmap-Duan-2021, clusnet-Yang-2019, dmap-Li-2020}, using them in semi-supervised settings with mean-teacher method \cite{mean_teacher-Antti-2018} is not straightforward. The external module may need additional loss functions, and often times they are trained before the detector with sufficient labeled data. Also, it is not immediately clear how to construct pseudo-labels for the density extraction module if one wants to train them in the mean-teacher settings using unlabeled images. Thus, in our case, we want to identify the density crops from the detector's prediction itself. For this, we used a cascade zoom-in detector (CZ Detector) design that learns to predict density crops as an additional class in addition to the base class objects. The CZ detector was introduced in \cite{czdet-akhil-2023}. In this work, we extend the CZ detector in order to exploit additional unsupervised data present in a semi-supervised setting. We show that a simple adaptation of the CZ detector for semi-supervised learning would not work and that some additional care is needed. Thus, experiments on two different datasets and meaningful ablation are reported that show the advantages of the proposed approach.

With the CZ detector, density crops can be identified on the labeled and unlabeled images. For the labeled images, they are identified a priori with the available ground-truth (GT) labels. For the unlabeled images, pseudo-GT predictions are utilized to locate the cluster of small objects and then labeled as density crops. The crops identified on both labeled and unlabeled images are used to augment the training set. The augmented crops result in more samples of small objects seen at higher pixel resolution, improving their detection chance. This is due to more pseudo-labels created on the unlabeled images (figure \ref{fig:pseudo_gt_ap_over_iter} right) compared to the vanilla mean-teacher method. The detector is then trained in the mean-teacher fashion with weak-strong augmentation consistency and pseudo labels for the unlabeled images. At inference, detection is performed separately on both the input image and upscaled density crops if any are present in that image. They are then fused and post-processed to get the final results. As the training and inference with crops here are relying only on the detector itself, their semi-supervised adaptation becomes easier. Figure \ref{fig:pseudo_gt_ap_over_iter} left shows how density crops are improving AP detection on the VisDrone dataset. It can be observed that by utilizing the density crops effectively in the semi-supervised settings, our detection accuracy increases significantly over the vanilla semi-supervised detector, as in the fully supervised settings.

Our main contributions can be summarized as follows:

\noindent \textbf{(1)} A density-crop guided semi-supervised detection method is proposed for aerial images. It adapts the vanilla mean-teacher semi-supervised detector with the capability to identify and process the cluster of small objects, improving their suitability for training semi-supervised detectors on high-resolution aerial images.

\noindent \textbf{(2)} The detector is designed in a zoom-in fashion that can identify clusters of small objects and re-detect the small objects by upscaling the clusters. Different from other approaches, the zoom-in operation doesn't need a separate module, the detector itself can identify the regions to zoom in. 

\noindent \textbf{(3)} We empirically validate the benefits of our semi-supervised detection method on aerial images from drones (VisDrone) and satellites (DOTA), and observed a consistent improvement in the detection accuracy on both datasets over the supervised training.

% !TEX root=main.tex

\section{Related works}
\label{sec:related-work}

%\subsection{Object Detection}
\noindent \textbf{Object Detection.}  
Generic object detectors have achieved impressive progress in recent years on object detection tasks in natural images \cite{faster_rcnn-Ren-2015, fpn-Lin-2017, yolo_9000-Redmon-2017, ssd-Liu-2016} object detectors. This has resulted in wider adoption of them in practical applications. With aerial images, Faster RCNN based detectors are widely used as they tend to be more accurate due to potential object region extraction performed in the first stage \cite{clusnet-Yang-2019, glsan-Deng-2020, dmap-Li-2020}. Recently, one-stage detectors are being explored in aerial images~\cite{querydet-Yang-2022}. These days, anchor-free detectors~\cite{fcos-Tian-2019, centernet-Duan-2019} are getting popular, since they avoid the need for hand-crafted anchor box dimensions and their matching process. Anchor-based matching is difficult with small objects as the overlap values are low for small bounding boxes. Though appealing, the anchor-free detectors are still challenging to use in semi-supervised settings as they produce many noisy pseudo labels \cite{dsl-Chen-2022}. This might result in substantial modifications to the backbone detector and additional loss functions. To avoid further challenges in terms of adapting the backbone detector, in this paper, we used the Faster RCNN detector which has shown excellent results in the existing works \cite{clusnet-Yang-2019, dmap-Duan-2021, glsan-Deng-2020}.

\noindent \textbf{Semi-supervised Object Detection}
Semi-supervised object detectors are trained on a small set of labeled images and a large collection of unlabeled images \cite{humble-teacher-Yang-2021}. They have been explored in many different forms. The dominant approaches for semi-supervised object detection in the past were consistency regularization or pseudo-label based\cite{stac-sohn-2015, soft-teacher-Xu-2021, csd-Jeong-2019}. But recently, the mean-teacher method became very popular in object detection which combines both approaches, resulting in state-of-the-art results\cite{unbiased-teacher-Liu-2021, humble-teacher-Yang-2021, soft-teacher-Xu-2021, class-prototype-Li-2022, sed-Guo-2022, stac-sohn-2015}. In this setting, a student network is trained to optimize the combined supervised and unsupervised loss. The supervised loss is the standard detection loss. The unsupervised loss forces the student network to make consistent predictions with a weak and strong augmented version of an unlabeled image. It is implemented by using pseudo labels from a teacher network which temporally accumulates the student network weights in an exponential moving average (EMA) fashion. Initially successful in two-stage detectors, the mean-teacher method is now getting popular with one-stage detectors too \cite{ubteacher_v2-Liu-2022, dsl-Chen-2022}. While existing semi-supervised detectors adapt the mean-teacher method to fit different requirements, we focus on adapting the mean-teacher detector to the small object detection setting. To this end, we propose a density crop-guided semi-supervised detector.

\noindent \textbf{Detection of Small Objects.} 
Small object detection is getting popular these days with the large-scale availability of drone and satellite images. Cheng et al. \cite{sod_survey-Cheng-2022} provides a comprehensive survey of small object detection techniques which is broadly classified into: scale aware training \cite{trident_nw-Li-2019, fpn-Lin-2017, snip-Singh-2018, sniper-Singh-2018}, super-resolution methods \cite{pgan-Li-2017}, context modeling \cite{inside_out_net-Bell-2016}, and density guided detection \cite{dmap-Duan-2021, dmap-Li-2020, clusnet-Yang-2019} among others. In aerial images, the small objects are distributed in clusters at sparse locations, so density cropping based approaches have shown excellent results \cite{dmap-Duan-2021, clusnet-Yang-2019, dmap-Li-2020, glsan-Deng-2020}. Thus we also used density crop-guided detection for improving the small object detection in this work. Different from others, we trained the density-based detector in semi-supervised settings. As most of the density-based detectors are trained with additional density extraction modules and loss functions, they cannot fit in the mean-teacher semi-supervised framework. So we used a density-based approach that performs density crop extraction within the detector itself. By doing so, we can easily adapt the mean-teacher semi-supervised training to incorporate small object detection guided by object density.
% !TEX root=main.tex
\section{Proposed Approach}
\label{sec:proposal}
In this section, we discuss the formulation of our semi-supervised zoom-in detection in detail. First, we present how we leverage the density crops for small object detection within the object detector using the zoom-in design. This design avoids the need for additional modules to get density crops simplifying the subsequent semi-supervised detector learning.  Then we present the semi-supervised detection in detail with the teacher-student learning paradigm (mean-teacher method). We present how density crops are used in semi-supervised settings where crops are identified on labeled and unlabeled images. While for the labeled images, the original GT annotation can be used, for the unlabeled images, we rely on the detection prediction. As the majority of the images are unlabeled in the standard semi-supervised settings, without using them we'll be limited to a few samples for the crop class. We utilized the pseudo-GT predictions on the unlabeled images to identify the dense cluster of small objects. Then the detector is trained with augmented crops from labeled and unlabeled images in a teacher-student mutual learning fashion. 

\begin{figure*}[t]
  \centering
  \includegraphics[height=7.8cm, width=0.9\textwidth]{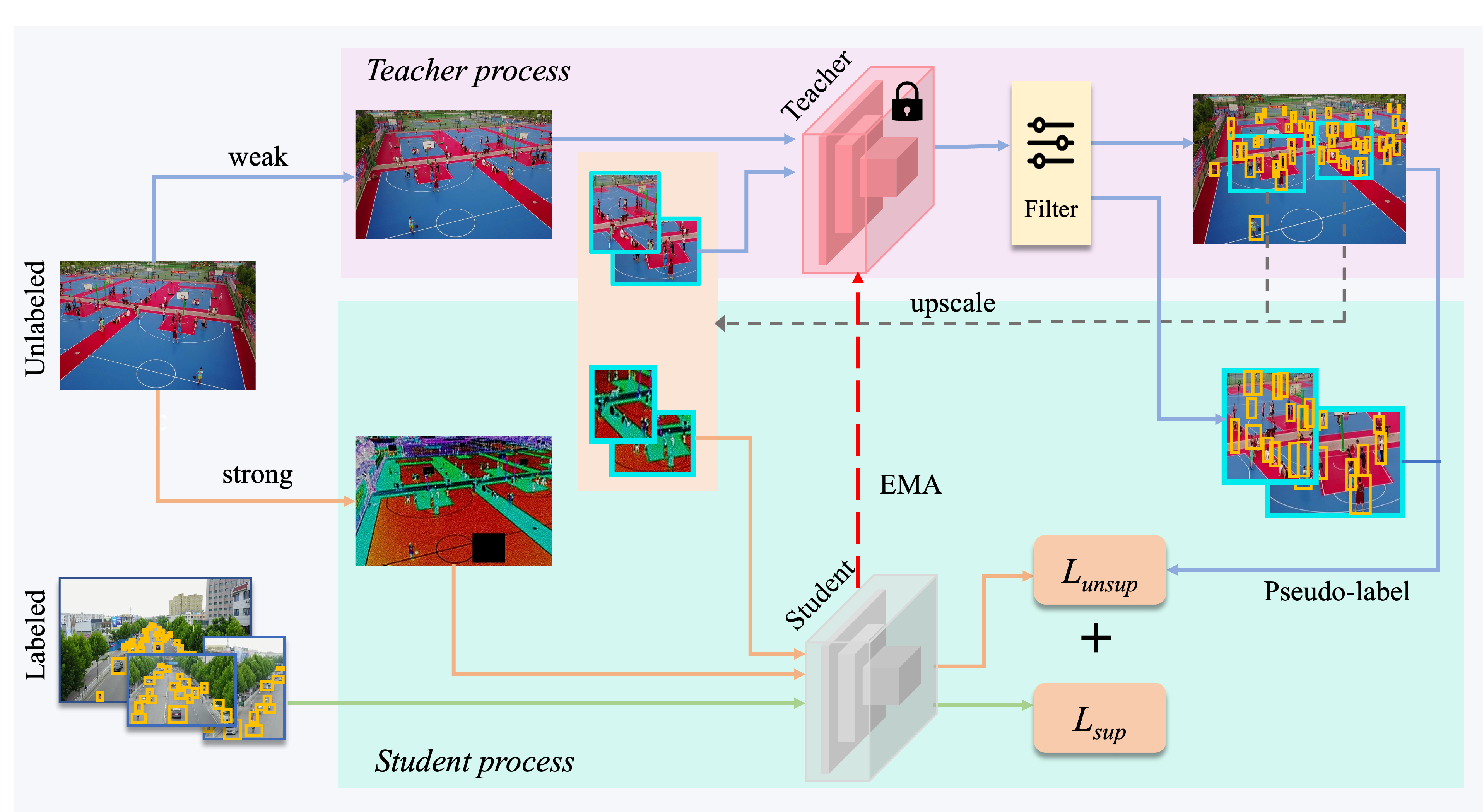} 
  \caption{The pipeline of our proposed density crop guided semi-supervised detection. The training data contains both labeled and unlabeled images. There are two networks that are identical copies of the backbone detector. The student network is learned via backpropagating the loss gradients, whereas the teacher network is an exponential moving average (EMA) of the student weights. The labeled images are passed through the student network and supervised loss $\mathcal{L}_{sup}$ is calculated. Unlabeled images are passed to the teacher network, whose predictions are then filtered (we used confidence thresholding here) to get good-quality pseudo-labels. If there are dense clusters of small objects in the unlabeled image, such clusters are cropped and passed after up-scaling to the teacher network. Then pseudo-labels are computed on newly added density crops as well in a similar fashion. A strongly augmented version of the unlabeled images and their density crops are then passed to the student network. The loss $\mathcal{L}_{unsup}$ is calculated based on the pseudo-labels obtained before. The combined loss is then  backpropagated to update the student weights. Teacher weights are then updated by EMA of the student weights.}
  \label{fig:semi_sup_system}
\end{figure*}

\subsection{Zoom-in Detector}
The zoom-in object detector helps us to zoom in on dense image regions containing cluster of small objects and detect them with better localization accuracy. It does the zoom-in operation like detecting any other class in the dataset, allowing us to use the detector off the shelf. - In this section, we will explain the working of zoom-in detection in detail. The first component of the zoom-in detector is the density crop labeling algorithm that labels the dense image regions as ``density crops'' and augments the training data by adding up-scaled versions of those crops. For the labeled data, we use the available GT boxes for the crop labeling. For the unlabeled data, we rely on pseudo labels for crop labeling. Then, the density crops are  added to the set of target objects to be detected on each image. The zoom-in detector is then trained in the mean-teacher fashion. The inference is performed in two stages. In the first stage, we detect base class objects and ``dense crops'' on the input image. In the second stage, the obtained dense image regions are up-scaled and a second detection is performed on them. This will help us to detect those small objects in the crowded image regions. A fusion operation is then performed by merging the detections on the input image and its dense regions which will be explained later.

\subsection{Density Crop Labeling}
To label density crops, we need to locate dense image regions containing cluster of small objects and group them in a single bounding box. These bounding boxes are then added to the training targets as a new class called ``density crop''. To locate the density crops, we identified the cluster of small objects using the algorithm \ref{alg:crop_discovery}. It basically computes the pairwise IoU(Intersection over Union) among the GT boxes and adjacent boxes are labeled as connected if there is an overlap above a threshold $\theta$. All those connected boxes are merged into a single large box (by finding an enclosing box based on the min and max coordinates of all connected boxes) identifying a dense image region. The merging and connection labeling operations are performed iteratively. Algorithm \ref{alg:crop_discovery} describes the procedures in detail.

\begin{algorithm}[t]
    \caption{Density Crop Labeling Algorithm.}
    \label{alg:crop_discovery}
    \KwIn{$\mathcal{B}$: GT boxes in an image}
    \KwOut{$\mathcal{D}$: Density crops}
    \Parameter{$N$: no. of merging steps, \\${\sigma}$: expansion pixels, \\${\theta}$: overlap threshold, \\${\pi}$: maximum crop size}
    1. $\mathcal{D}  \leftarrow$ {\texttt{scale}}($\mathcal{B},{\sigma}$)\;
    2. \For{$i\gets1$ \KwTo $N$ }{
        %a) $O$ = pairwise\_IOU($\mathcal{B}^*$) \#overlap matrix \\%($|\mathcal{C}| \times |\mathcal{C}|$)\\
        %b) $C$ = $O$ $>$ {\color{red} $\theta$} \ \ \#connection matrix \\%$(K \times K)$ \\
        %a) $O$ = pairwise\_connection($\mathcal{B}^*,\theta$) \\%\#connection matrix \\%
        a) $O$ = {\texttt{pairwise\_IoU}}($\mathcal{D}$)\\
        b) $C = O > \theta$ \\
        c) $\mathcal{D} \leftarrow \emptyset$\\ 
        \While{$|C|>0$}{
        i) $m^* = \text{arg} \max_{m} \left( \sum_i C_{m,i} \right)$ \\ %\#max connections \\
        ii) $d$ = {\texttt{enclosing\_box}}($C_{m^*}$) \\%\#new density crop \\        
        %iii) $\mathcal{B}^* \leftarrow$ $\mathcal{B}^*-link(C_m)$\\
        %iv) $C \leftarrow$ $C - link(C_m)$\\
        iii) $\mathcal{D} \leftarrow$ $\mathcal{D}+ d $ \\
        iv) $C_{m^*}$ = 0 \\%\vec{0}$\\
        }
        d) $\mathcal{D} \leftarrow$ {\texttt{filter\_size}}$(\mathcal{D},\pi)$\\
    }
\end{algorithm}

First, all GT boxes $\mathcal{B}$ are scaled by expanding their min and max coordinates by  $\sigma$ pixels (\verb|scale|($\mathcal{B},\sigma$)). Then the pairwise IoU between the scaled boxes (\verb|pairwise_IoU|($\mathcal{D}$)) is calculated in $O$ as a $|\mathcal{D}| \times|\mathcal{D}|$ matrix. Two boxes are deemed as connected if their overlap is above a threshold $\theta$. $C$ matrix stores all the pairwise connections. Then we select the box $m^*$ with the maximum number of connections from $C$. An enclosing box is computed (\verb|enclosing_box|($C_{m^{*}}$)) by finding the min and max coordinates of all boxes connected to $m^*$. The newly obtained box is added to the list of crops. Then all connections from the box $m^*$ are removed by setting the row $C_{m^*}$ to zero. Thus we complete the discovery of one density crop. This procedure is repeated until all the connections are removed finding many dense regions in the image. Then we perform a filtering operation where crops bigger than a maximum threshold $\pi$ are removed from the list $\mathcal{D}$ (\verb|filter_size|($\mathcal{D},\pi$)). This concludes one step of iterative merging. We perform this operation $N$ times by considering the newly discovered crops as GT boxes. Without the iterative merging, there will be many redundant crops as observed in \cite{clusnet-Yang-2019}. The size threshold $\pi$ for the crops is the ratio of the area of the crop to that of the image. 

\begin{figure}
\centering
\begin{tabular}{c  @{\hspace{-0.3cm}} c}
\includegraphics[width=0.51\linewidth, height=2.8cm]{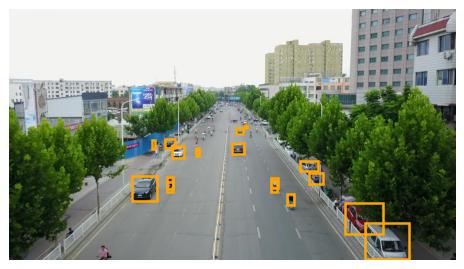} &
\includegraphics[width=0.51\linewidth, height=2.8cm]{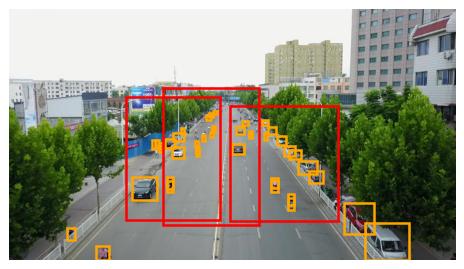}\\
(a) basic detector & (b) zoom-in detector \\
\end{tabular}
 \caption{Detection using the zoom-in detector. (a) detection with a standard Faster RCNN. (b) detection with the zoom-in Faster RCNN. The density crops are shown in red color. The zoom-in detector is detecting many small objects inside the crop regions.}
 \label{fig:clusters}
\end{figure}

The newly obtained crops are also used for augmenting the training set. During training time, the detector is processing 
an up-scaled version of the density crops, and the loss is computed against the target small objects in the crop. As we have the small objects in higher pixel resolution in the up-scaled version, the detector can recognize them easily. Note that in the original input where we resize the image to the detector's maximum training resolution, the small objects from these crowded dense regions are mostly missed by the detector. But when we process the up-scaled density crops, we improve the chance of detecting them. See figure \ref{fig:clusters} for an illustration of the zoom-in detection in action.

\subsection{Semi-supervised Training}
Semi-supervised learning takes place by distilling the weights of a detector (called a student network) during training to another identical copy of the network (called the teacher network) by exponential moving average (EMA). The teacher network is generally more stable due to the slower pace at which it temporally ensembles the noisy student weights, so it is used to give pseudo GT for the unlabeled images\cite{unbiased-teacher-Liu-2021}. The student network learns its weights by optimizing a combination of supervised and unsupervised loss. For the labeled data, we have the GT annotations to compute the supervised loss $\mathcal{L}_{sup}$. Let the available labeled data is $D_s = \{x_i, y_i\}_{i=1}^{N_s}$, where each $y_i$ is a bounding box coordinate and its class label ($y_i = (b_i, c_i)$). Here ${N_s}$ is the number of labeled samples. For the unlabeled data $D_u = \{x_i\}_{i=1}^{N_u}$, we get pseudo GT $\hat{y}_i$ from the teacher network which is used to calculate the unsupervised loss $\mathcal{L}_{unsup}$. Here ${N_u}$ is the number of unlabeled samples. Finally, the network is trained by optimizing the following loss
\begin{equation}
    \mathcal{L} = \mathcal{L}_{sup} + \lambda \mathcal{L}_{unsup}
    \label{eq:overall_loss}
\end{equation}
where $\lambda$ is a hyperparameter to control the relative importance of the supervised and unsupervised loss. Figure \ref{fig:semi_sup_system} shows the overall architecture of our semi-supervised learning system. At each iteration, we sample a minibatch of labeled and unlabeled samples following a preset ratio $d_r$. Each datapoint in the minibatch undergoes two types of transformation, referred to as weak and strong augmentation. The weak augmentation is simply the rescaling and horizontal flip transformation. The strong augmentation includes color jittering, grayscale, Gaussian blur, and cutout patches which perform only pixel-level transforms, thus the bounding box labels need not be transformed. We followed the scale ranges provided in \cite{unbiased-teacher-Liu-2021} for the strong augmentation. The augmented images then go through the mean-teacher semi-supervised learning process. We followed \cite{unbiased-teacher-Liu-2021} for the mean-teacher training implementation. We compute density crops on the unlabeled images using pseudo-labels from the teacher. This is then used to augment more crops, this time from the unlabeled images. In the following, we will describe the semi-supervised learning process in detail.

\subsubsection{Burn-in stage}
To get reliable pseudo GT for the unlabeled images, the teacher network should have a good initialization. Typically, existing methods perform a supervised pre-training with the available supervised data to get this good initialization\cite{stac-sohn-2015, humble-teacher-Yang-2021, unbiased-teacher-Liu-2021}. This supervised pre-training is called Burn-in stage. During burn-in, we optimize $\mathcal{L}_{sup}$ only which is a sum of classification and localization losses of the detector.

\begin{equation}
    \mathcal{L}_{sup} = \sum_{i=1}^{N_s}\mathcal{L}_{cls}(f_{W}(x_i), y_i) + \mathcal{L}_{reg}(f_{W}(x_i), y_i)
    \label{eq:sup_loss}
\end{equation}

After burn-in, the weights of the network $W$ are copied to the teacher ($W \rightarrow W_t$) and student network ($W \rightarrow W_s$). From this point, unsupervised data is also used in the learning process with teacher-student mutual learning.

\subsubsection{Teacher-student learning stage}
The teacher-student learning process optimizes the loss in equation \ref{eq:overall_loss} to learn the student network (with backpropagation), whereas the teacher network is learned by temporally accumulating the student weights (with EMA). It combines consistency regularization and pseudo label-based learning - the most popular approaches for semi-supervised learning - in one framework. The consistency regularization is ensured with the weak-strong augmentation prediction consistency. Pseudo label based learning is performed by producing pseudo labels on the unlabeled images. 

The weakly augmented version of unlabeled data first goes through the teacher network producing the instance predictions. This prediction then undergoes confidence thresholding to produce pseudo labels $\hat{y}$. Let $y^{pred}_j = (b^{pred}_j, c^{pred}_j, p^{pred}_j)$ are instance predictions containing predicted box $b^{pred}_j$, class $c^{pred}_j$ and probability $p^{pred}_j$ where $y_{pred}$ is obtained as
\begin{equation}
y^{pred} = f_{W_t}(x)
\end{equation}
The confidence thresholding considers all predictions with  a class probability above a threshold $\tau$ as foreground instances:
\begin{equation}
    \hat{y} = \{y^{pred}_j | p^{pred}_j > \tau, \forall j \in y^{pred} \}
    \label{eq:pseudo_label}
\end{equation}
This is the filtering process shown in figure \ref{fig:semi_sup_system}. Once we get the pseudo labels for the unlabeled images, we can compute the $\mathcal{L}_{unsup}$. For that, the strongly augmented version of the unlabeled data is passed through the student network to get the predictions. The unsupervised loss is then applied to the classification head as follows: 
\begin{equation}
    \mathcal{L}_{unsup} = \sum_{i=1}^{N_u}\mathcal{L}_{cls}(f_{W_s}(x_i), \hat{y}_i)
    \label{eq:unsup_loss}
\end{equation}
$\mathcal{L}_{unsup}$ is not applied to the localization head of the detector because the confidence thresholding based pseudo labeling is suitable only for getting confident class predictions, it has no information about the bounding box correctness. After computing $\mathcal{L}_{unsup}$, we update the student network weights $W_s$ by optimizing equation \ref{eq:overall_loss}. The teacher weights $W_t$ is then updated by EMA as follows:
\begin{equation}
    W_t = \alpha W_t + (1-\alpha) W_s
    \label{eq:ema_update}
\end{equation}
where $\alpha$ is a hyperparameter that controls the pace at which the student weights are updated to the teacher weights.

\subsection{Density Crops on Unlabeled Images}
As density crops help to process crowded image regions in higher pixel resolution and improve small object detection performance, it is useful to find them on unlabeled images as well. While for the labeled data $D_s$ we have the GT labels $y$ to run crop-labeling algorithm \ref{alg:crop_discovery}, we don't have annotations for  the unlabeled data $D_u$ to produce density crops. As we have plenty of unlabeled images, we could get more augmented crops from dense regions of unlabeled images, also increasing samples for the crop category. Thus we expect further improvement in performance if density crop-based learning can be utilized on unlabeled images as well. To do so, we rely on the predictions of the teacher network. Particularly, we utilize the pseudo labels provided by the teacher network to label crops on the unlabeled images, again using algorithm \ref{alg:crop_discovery}.

After the semi-supervised training with labeled and unlabeled data (where crops are only augmented on labeled images) is converged, we use the final teacher model to get the predictions on the unlabeled images. These predictions are then processed to get accurate pseudo GTs following confidence thresholding as in equation \ref{eq:pseudo_label}. Crop labeling on the unlabeled images is then performed following algorithm \ref{alg:crop_discovery} this time with pseudo GT boxes. The semi-supervised training is then continued as before but with more unlabeled datapoints obtained from the cluster of small objects in the unlabeled images. As the clusters mostly remain intact on the unlabeled images at this point, it is not necessary to recompute them at every iteration. We recomputed them at every 10,000 iterations to make the training faster. 

\begin{figure}[t]
  \centering
  \includegraphics[height=3.7cm, width=0.48\textwidth]{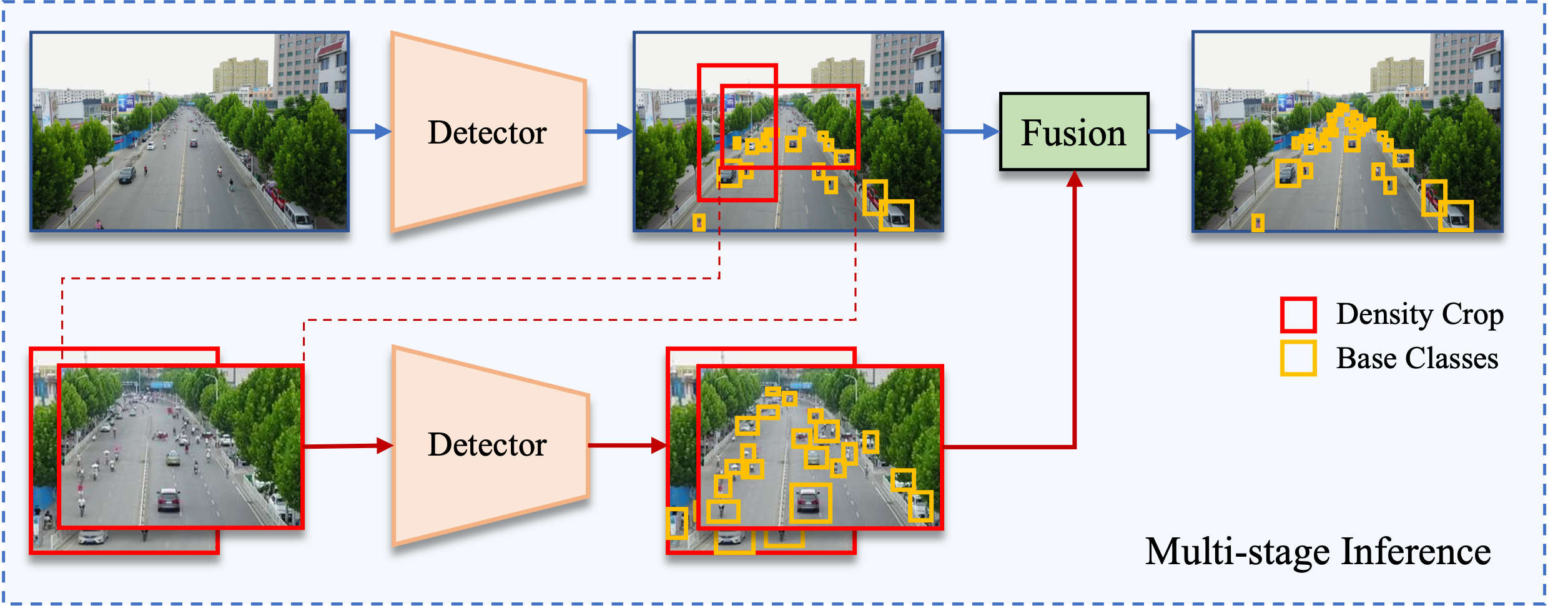} 
  \caption{Multi-stage inference with density crops. During the first stage, detections are obtained on the input image. Density crops are then derived using these detections. In the second stage, the density crops are upscaled and fed to the detector again followed by a second inference. Finally, the detections on density crops are combined with the detections on the input image.}
  \label{fig:dcrop_inf}
\end{figure}

\subsection{Multi-stage Inference}
The density crops are leveraged at inference by performing detection on the up-scaled crops and fusing that detection with the detection on the input image. In this way, the crowded small object regions are processed at a higher scale, facilitating better small object detection. Figure~\ref{fig:dcrop_inf} bottom explains our inference process in detail. It consists of two stages. In stage one, it predicts detections on the input image. Then we can obtain density crops from this detection in two ways. 
\begin{enumerate}
    \item select the high-quality density crops based on their confidence score of the predicted crop instances.
    \item perform crop labeling with the predicted confident detections to get density crops.
\end{enumerate}
The empirical studies show that predicted crops are faster at inference than labeled crops. But, labeled crops are more accurate than the predicted ones. So, depending on the speed vs accuracy trade-off of the downstream application, one can select the suitable inference procedure. In stage two, the upscaled density crops are passed through the same detector again, producing small object detection on the density crops. Finally, we re-project the detections on the crops to the original image and concatenate them with the detections on the original image. 
Let $d \in \mathcal{D}$ be an up-scaled crop image of size $(I^W_d,I^H_d)$ defined by its bounding box coordinates $(d_{x1}, d_{y1}, d_{x2}, d_{y2})$ in the original image. Given the scaling factors $(S^W_d,S^H_d)=(\frac{d_{x2}-d_{x1}}{I^W_d},\frac{d_{y2}-d_{y1}}{I^H_d})$, the re-projection box $p_i$ scales down and shifts the predicted boxes  $(x_{1,i},y_{1,i},x_{2,i},y_{2,i}) \in \mathcal{B}^d$ in the crop $d$ as:
\begin{align}
    p_i =  & ({S^W}x_{1,i},{S^H}y_{1,i},{S^W}x_{2,i},{S^H}y_{2,i}) \notag \\
    & + (d_{x1}, d_{y1}, d_{x1}, d_{y1})
    \label{equ:reproj}
\end{align}

The Non-Maximal Suppression (NMS) is then applied to remove duplicate detections. All of these operations can be easily wrapped on top of a detector's inference procedure without changing its internal operations.
%%%%%%%%
\subsection{Semi-supervised Training Algorithm}
Algorithm \ref{alg:train_algo} summarizes our density crop-guided semi-supervised training process. Given the labeled and unlabeled data $D_s$ and $D_u$ respectively, we first compute and label crops in $D_s$ using the available ground-truth labels. The training process then begins. We load a batch of images from both the labeled and unlabeled pool. The batch loaded from labeled pool $x_s$ is directly used to calculate $\mathcal{L}_{sup}$. For the batch from unlabeled pool $x_u$, strongly augmented and weakly augmented versions are produced. The teacher processes weakly augmented images computing pseudo labels $\hat{y}_u$ for the images in $x_u$. This is then used to compute $\mathcal{L}_{unsup}$ where the loss is computed against the student predictions obtained using strongly augmented images. The combined loss $\mathcal{L}$ is backpropagated, and then teacher weights are updated using the EMA update rule in equation \ref{eq:ema_update}. When this training process is converged (after a sufficient number of iterations $n$), crops are computed on the unlabeled images and used to further augment $D_u$.

\begin{algorithm}[t]
    \caption{Density-crop Semi-supervised Training}
    \label{alg:train_algo}
    \KwIn{labeled and unlabeled images: $D_s, D_u$}
    \KwOut{teacher and student weights: $W_t, W_s$}
    \Parameter{$n$: start of labelling $D_u$ iteration, \\$N$: maximum iterations, $\mu$: learning rate}
    1. Label and augment crops in $D_s$ using algorithm \ref{alg:crop_discovery}\\%using available annotations $\mathcal{B}$.\\
    %2. $DL_s, DL_u \leftarrow$ data\_loader($D_s$), data\_loader($D_u$)
    3. \For{$i\gets1$ \KwTo $N$ }{
        a) $x_s, y$ = batch($D_s$) \\
        b) $x_u$ = batch($D_u$)\\
        b) Compute $\mathcal{L}_{sup}(x_s,y)$ using eqn. \ref{eq:sup_loss}\\
        c) Obtain $\hat{y}_u$ using eqn. \ref{eq:pseudo_label}\\
        d) Compute $\mathcal{L}_{unsup}(x_u, \hat{y}_u)$ using eqn. \ref{eq:unsup_loss}\\
        e) Compute $\mathcal{L}$ using eqn. \ref{eq:overall_loss}\\
        f) $W_s \leftarrow W_s - \mu \frac{\partial \mathcal{L}}{\partial W_s}$\\
        g) update $W_t$ using eqn. \ref{eq:ema_update}\\
        i) \If{i $\geq$ n and $x_u$ is not augmented}{
         I) Label and augment crops in $D_u$ for the batch $d_u$ using pseudo ground-truth $\hat{y}_u$.\\
         II) set $d_u$ augmented to True.
        }
    }
\end{algorithm}

% !TEX root=main.tex

\section{Experiments}
\label{sec:experiments}

%%
%\subsection{Experimental Setup}
\noindent \textbf{Datasets and evaluation measures.} For the evaluation of methods, we employed two popular challenging benchmark datasets for Aerial Image Detection, namely the VisDrone \cite{zhu-VisDrone-2018} and DOTA \cite{xia-DOTA-2018} datasets. We used COCO style AP for assessing and comparing the performance of the methods \cite{mscoco-Lin-2014}. The AP of small, medium, and large objects are also reported, in particular, to understand the performance of methods for small-scale object detection in aerial images. 

\noindent \textbf{VisDrone.} This dataset contains  8,599 drone-captured images (6,471 for training, 548 for validation, and 1,580 for testing) with a resolution of about 2000 $\times$1500 pixels. The objects are from ten categories with 540k instances annotated in the training set, mostly containing different categories of vehicles and pedestrians observed when the drone is flying through the streets. It has an extreme class imbalance and scale imbalance making it an ideal benchmark for studying small object detection problems. As the evaluation server is closed now, following the existing works, we used the validation set for evaluating the performance.

\noindent \textbf{DOTA.} This dataset is comprised of satellite images. The images in this dataset have a resolution ranging from 800$\times$800 to 4000$\times$4000. Around 280k annotated instances are present in the dataset. The objects are from fifteen different categories, with movable objects such as planes, ships, large vehicles, small vehicles, and helicopters. The remaining ten categories are roundabouts, harbors, swimming pools, etc. The training and validation data contain 1411 images and 458 images, respectively. Since the images of DOTA are too large to be fed to the network directly, we extracted 1500$\times$1500 crops from the image by shifting 1000 pixels in a sliding window fashion.

\noindent \textbf{Implementation details.} The Detectron2 toolkit \cite{detectron2-wu-2019} was used to implement our CZ detector. The backbone detector used in our study is Faster RCNN \cite{faster_rcnn-Ren-2015}. We used Feature Pyramid Network (FPN) \cite{fpn-Lin-2017} backbone with ResNet50 \cite{resnet-He-2016}  pre-trained on ImageNet \cite{imagenet-Russakovsky-2015} dataset for our experimental validation. For data augmentation, we resized the shorter edge to one randomly picked from (800, 900, 1000, 1100, 1200), and applied horizontal flip with a 50\% probability. The model was trained on both datasets for 180k iterations. The initial learning rate is set to 0.01 and decayed by 10 at 70k iteration. For training, we used one NVIDIA A100 GPU with 40 GB of memory.

%%%%%%%%
\subsection{Comparison with Different Percentage of Labeled Data}
We analyzed the effectiveness of our semi-supervised learning method by using partially labeled data from the train set of VisDrone and DOTA datasets. In particular, we used 1\%, 5\%, and 10\% randomly chosen data points from the train set as labeled data and the remaining as unlabeled for the semi-supervised training. There are five settings in the comparison; supervised baseline, supervised baseline with density crops (Supervised + Crop), semi-supervised with the mean teacher (SSOD), SSOD with density crops on labeled images (SSOD + Crop (L)), and SSOD with density crops on labeled and unlabeled images (SSOD + Crop (L + U)). These settings progressively assess the impact of the components of our density crop-guided semi-supervised object detection.

Table \ref{table:diff_sup_data_visdrone} presents the results for the VisDrone\cite{zhu-VisDrone-2018} dataset.  It compares the detection average precision values obtained using the COCO evaluation protocol \cite{mscoco-Lin-2014} for Intersection over Union (IoU) thresholds [0.5:0.05:0.95] (\textbf{AP}), and 0.5 (\textbf{$\textrm{AP}_{50}$}). It can be observed that \textbf{AP} is improved by more than 6\% in all cases with our density-guided SSOD over their supervised baseline. Compared to the vanilla mean-teacher method (SSOD), our density crop-guided SSOD shows an average improvement of more than 2\% on all metrics. Compared to 1\% and 5\% cases, with very limited labeled samples per class, 10\% shows a better boost in performance while leveraging density crops with SSOD. Another interesting result is that the improved performance with semi-supervised learning for 1\% settings is more than that of supervised training with 5\% labels and 2\% below with the 10\% labels. This is achieved with less than 100 labeled samples. $\textrm{AP}_{50}$ has a gain of more than 5\% compared to the vanilla mean-teacher when semi-supervised learning is performed with density crops in the 10\% setting.  

%FPS 29.08 for Base Detector, 15.13 for crop detector
\begin{table*}
    \caption{Performance comparison of our density crop guided semi-supervised object detection with 1\%, 5\%, and 10\% labeled images on the VisDrone dataset. SSOD - semi-supervised detection with mean-teacher, Crop(L) - density crops on the labeled images, Crop (L + U) - density crops on the labeled and unlabeled images.}
    \centering
    \resizebox{.97\textwidth}{!}{
    \begin{tabular}{l||cc|cc|cc}
    \hline
    \textbf{Settings} & \multicolumn{2}{|c|}{1\% (\#Labeled =64)} & \multicolumn{2}{|c|}{5\% (\#Labeled =323)} & \multicolumn{2}{c}{10\% (\#Labeled =647)}\\
    %\hline
     & \textbf{AP} & \textbf{$\textrm{AP}_{50}$} & \textbf{AP} & \textbf{$\textrm{AP}_{50}$} & \textbf{AP} & \textbf{$\textrm{AP}_{50}$} \\
     \hline \hline
     Supervised & 10.69$\pm0.20$ & 23.53$\pm0.17$ & 16.11$\pm0.14$ & 32.55$\pm0.16$ & 19.26$\pm0.13$ & 37.73$\pm0.22$ \\
     Supervised + Crop & 13.04$\pm0.23$ & 26.97$\pm0.16$ & 20.90$\pm0.28$ & 40.21$\pm0.29$ & 23.30$\pm0.17$ & 43.85$\pm0.31$ \\
     \hline
     SSOD & 15.30$\pm0.29$ & 29.16$\pm0.48$ & 21.89$\pm0.16$ & 40.58$\pm0.29$ & 24.39$\pm0.22$ & 43.05$\pm0.20$ \\
     SSOD + Crop (L) & 16.64$\pm0.24$ & 31.11$\pm0.18$ & 22.52$\pm0.12$ & 41.33$\pm0.24$ & 26.48$\pm0.19$ & 47.50$\pm0.14$ \\
     \rowcolor{mygray} SSOD + Crop (L + U) & \textbf{17.21}$\pm0.16$ & \textbf{31.22}$\pm0.21$ & \textbf{23.57}$\pm0.21$ & \textbf{42.34}$\pm0.19$ & \textbf{27.46}$\pm0.16$ & \textbf{48.95} $\pm0.14$ \\
     \hline  
    \end{tabular}
    }
    \label{table:diff_sup_data_visdrone}
\end{table*}

\begin{figure}[t]
  \centering
  \includegraphics[height=6cm, width=0.49\textwidth]{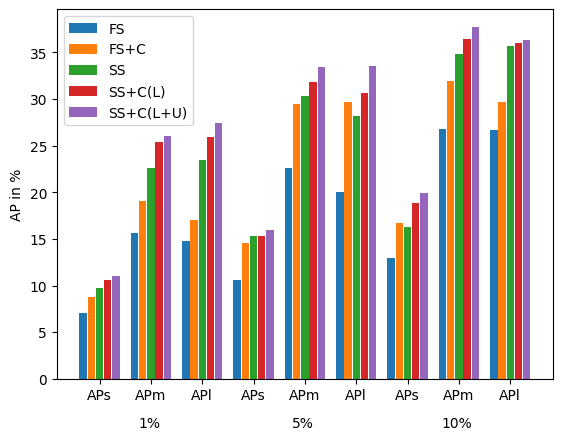} 
  \caption{Detection AP of small, medium, and large objects with different percentages of supervised data on the VisDrone dataset. FS: fully supervised, FS+C: fully supervised with crops, SS: vanilla mean-teacher, SS+C: mean-teacher with density crops on labeled images, SS+C+U: mean-teacher with density crops on all images.}
  \label{fig:vis_sml}
\end{figure}

We also studied how the AP of small, medium, and large objects behave in the same five settings described above. Figure \ref{fig:vis_sml} shows the results. The trend here is similar to that of table \ref{table:diff_sup_data_visdrone}. Using density crops increases the detection accuracy both in supervised and semi-supervised settings. Compared to the supervised settings, the AP of all-sized objects increases by more than 5\% when semi-supervised learning is performed with density crops. The improvement over the vanilla mean-teacher is more than 3\% in most settings. The APs of small, medium, and large objects with fully supervised training using 100\% labeled data are 25.74, 42.93, and 41.44 respectively. It can be observed that our model with 10\% labeled data performs competitively with this fully supervised upper bound.

\begin{figure}[t]
  \centering
  \includegraphics[height=6cm, width=0.49\textwidth]{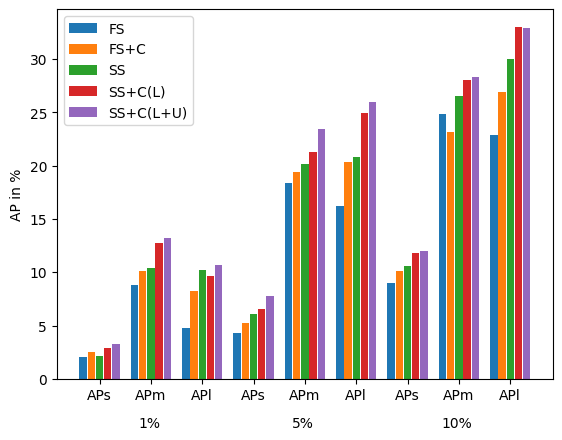} 
  \caption{Detection AP of small, medium, and large objects with different percentages of supervised data on the DOTA dataset. FS: fully supervised, FS+C: fully supervised with crops, SS: vanilla mean-teacher, SS+C: mean-teacher with density crops on labeled images, SS+C+U: mean-teacher with density crops on all images.}
  \label{fig:dota_sml}
\end{figure}

We further verified this observation  by conducting the same type of study in the satellite images of the DOTA dataset. Table \ref{table:diff_sup_data_dota} shows the results. The magnitude of improvements is comparable to that of the VisDrone dataset. AP shows an average improvement above 2\% compared to the mean-teacher method. $\textrm{AP}_{50}$ has a gain of more than 3\% in this dataset compared to the mean-teacher. Also, the APs of small, medium, and large objects are studied in the same way as above. Figure \ref{fig:dota_sml} shows the results. APs of small, medium, and large objects with 100\% supervised data on the DOTA dataset are 15.66, 38.16, and 44.2 respectively. While for small objects, our method with 10\% labeled data is 3\% below the supervised upper bound, the gap is around 10\% for medium and large objects. This implies the boost from the density-guided training is more concentrated on the small objects. All of these experiments confirm the impact of each component in our model as well. The performance gain with our density-guided semi-supervised detector over the supervised baseline is significant and consistent.
\begin{comment}
\begin{table*}
    \centering
    \begin{tabular}{l||rrr|rrr|rrr|r}
    \hline
    \textbf{Settings} & \multicolumn{3}{|c|}{1\% (\#L=64)} & \multicolumn{3}{|c|}{5\% (\#L=323)} & \multicolumn{3}{|c|}{10\% (\#L=647)} & FPS\\
    \hline
     & \textbf{AP} & \textbf{$\textrm{AP}_{50}$} & \textbf{$\textrm{AP}_{75}$} &  \textbf{AP} & \textbf{$\textrm{AP}_{50}$} & \textbf{$\textrm{AP}_{75}$} & \textbf{AP} & \textbf{$\textrm{AP}_{50}$} & \textbf{$\textrm{AP}_{75}$} &  \\
     \hline
     Supervised & 5.56 & 12.39 & 3.92 & 14.55 & 25.52 & 14.27 & 19.18 & 34.44 & 18.40 & 0.50 \\
     Supervised + Dcrop & 6.80 & 14.19 & 5.51 & 15.97 & 28.98 & 15.74 & 20.43 & 36.50 & 20.27 & 0.92 \\
     \hline
     SSOD & 8.78 & 16.18 & 8.25 & 16.78 & 29.59 & 16.11 & 23.15 & 39.49 & 23.86 & 0.50 \\
     SSOD + Dcrop (L) & 9.74 & 18.44 & 8.96 & 18.42 & 31.62 & 18.26 &  24.34 & 42.30 & 24.29 & 0.92 \\
     SSOD + Dcrop (L + U) & \textbf{10.32} & \textbf{19.70} & \textbf{9.33} & \textbf{19.96} & \textbf{34.78} & \textbf{19.65} & \textbf{25.17} & \textbf{43.07} & \textbf{24.57} & 0.92\\
    \hline  
    \end{tabular}
    \caption{Performance comparison of our density crop guided semi-supervised object detection with 1\%, 5\%, and 10\% labeled images on the DOTA dataset. The detection speed is also reported in FPS. SSOD - semi-supervised detection with mean-teacher, Dcrop(L) - density crops on the labeled images, Dcrop (L + U) - density crops on the labeled and unlabeled images.}
    \label{table:diff_sup_data_dota}
\end{table*}
%%%%%
\end{comment}

\begin{table*}
    \centering
    \caption{Performance comparison of our density crop guided semi-supervised object detection with 1\%, 5\%, and 10\% labeled images on the DOTA dataset. SSOD - semi-supervised detection with mean-teacher, Crop(L) - density crops on the labeled images, Crop (L + U) - density crops on the labeled and unlabeled images.}    
    \resizebox{.97\textwidth}{!}{
    \begin{tabular}{l||cc|cc|cc}
    \hline
    \textbf{Settings} & \multicolumn{2}{|c|}{1\% (\#Labeled=14)} & \multicolumn{2}{|c|}{5\% (\#Labeled=71)} & \multicolumn{2}{c}{10\% (\#Labeled =141)} \\
    \hline
     & \textbf{AP} & \textbf{$\textrm{AP}_{50}$} &  \textbf{AP} & \textbf{$\textrm{AP}_{50}$} &  \textbf{AP} & \textbf{$\textrm{AP}_{50}$} \\
     \hline
     Supervised & 5.56$\pm0.26$ & 12.39$\pm0.49$ & 14.55$\pm0.11$ & 25.52$\pm0.24$ & 19.18$\pm0.18$ & 34.44$\pm0.36$ \\
     Supervised + Crop & 6.80$\pm0.19$ & 14.19$\pm0.41$ & 15.97$\pm0.18$ & 28.98$\pm0.35$ & 20.43$\pm0.15$ & 36.50$\pm0.51$ \\
     \hline
     SSOD & 8.78$\pm0.21$ & 16.18$\pm0.40$ & 16.78$\pm0.13$ & 29.59$\pm0.23$ & 23.15$\pm0.14$ & 39.49$\pm0.29$ \\
     SSOD + Crop (L) & 9.74$\pm0.18$ & 18.44$\pm0.26$ & 18.42$\pm0.16$ & 31.62$\pm0.25$ &  24.34$\pm$0.07 & 42.30$\pm0.11$ \\
      \rowcolor{mygray} SSOD + Crop (L + U) & \textbf{10.32}$\pm0.17$ & \textbf{19.70}$\pm0.22$ & \textbf{19.96}$\pm0.14$ & \textbf{34.78}$\pm0.24$ & \textbf{25.17}$\pm0.11$ & \textbf{43.07}$\pm0.21$ \\
    \hline  
    \end{tabular}% <------ Don't forget this %
    }
    \label{table:diff_sup_data_dota}
\end{table*}

\begin{figure*}
\centering
\begin{tabular}{c c @{\hspace{-0.3cm}} c @{\hspace{-0.3cm}} c @{\hspace{-0.3cm}} c}
\rotatebox{90}{\quad \quad Sup. DOTA} & \includegraphics[width=0.25\linewidth, height=3.2cm]{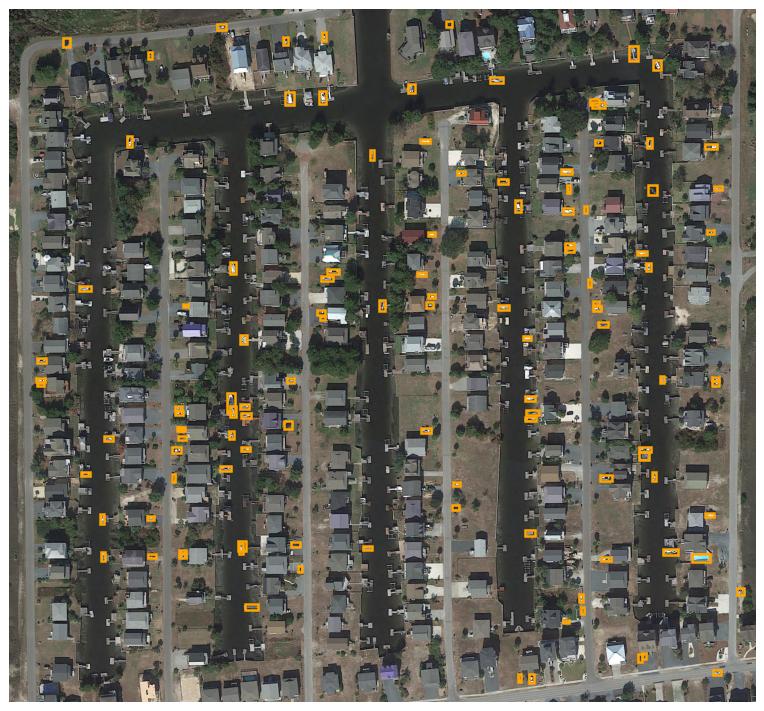} &
\includegraphics[width=0.25\linewidth, height=3.2cm]{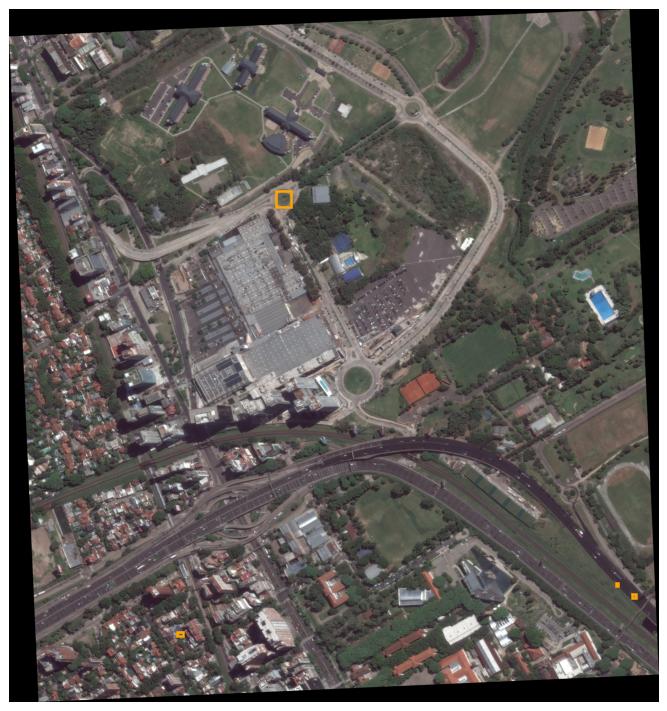} &
\includegraphics[width=0.25\linewidth, height=3.2cm]{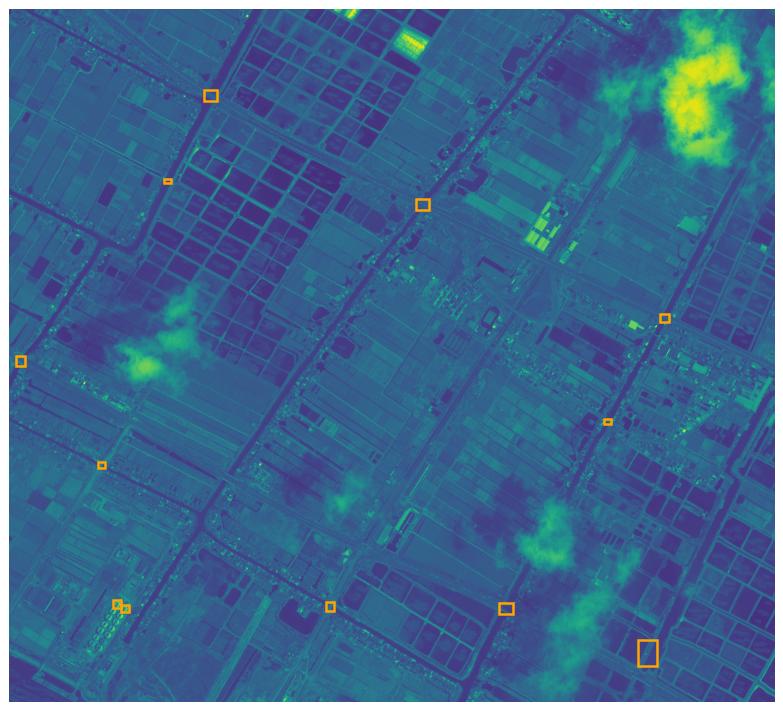} &
\includegraphics[width=0.25\linewidth, height=3.2cm]{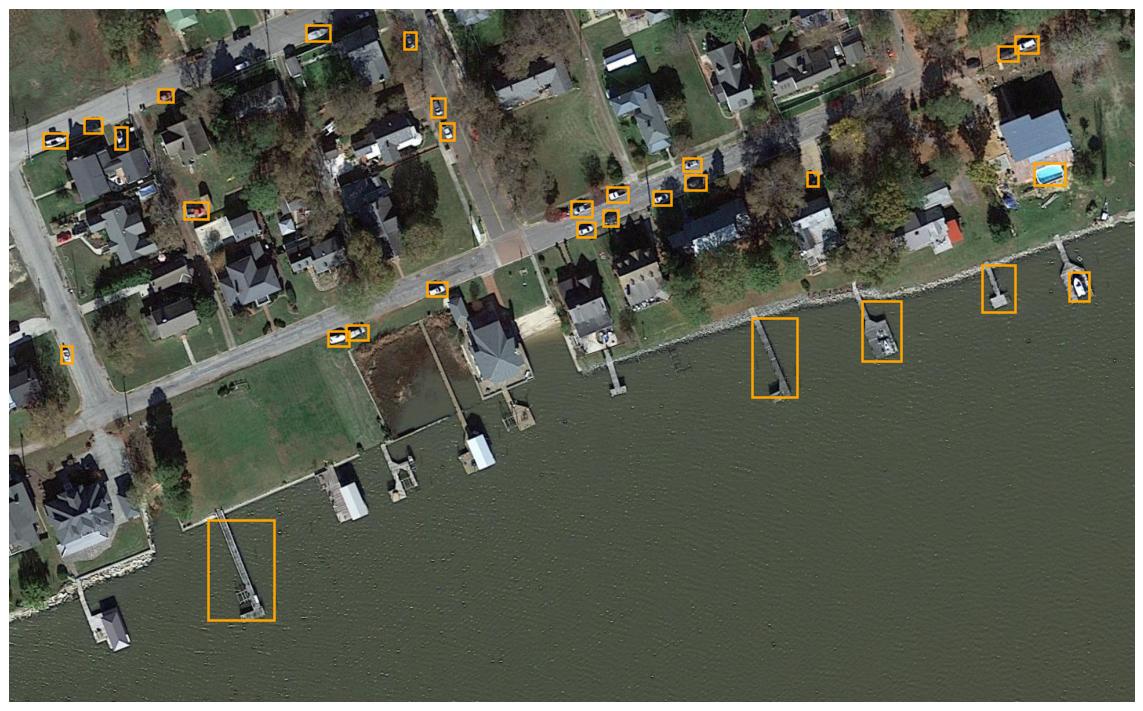}\\
\rotatebox{90}{\quad \quad SSOD+C DOTA} & \includegraphics[width=0.25\linewidth, height=3.2cm]{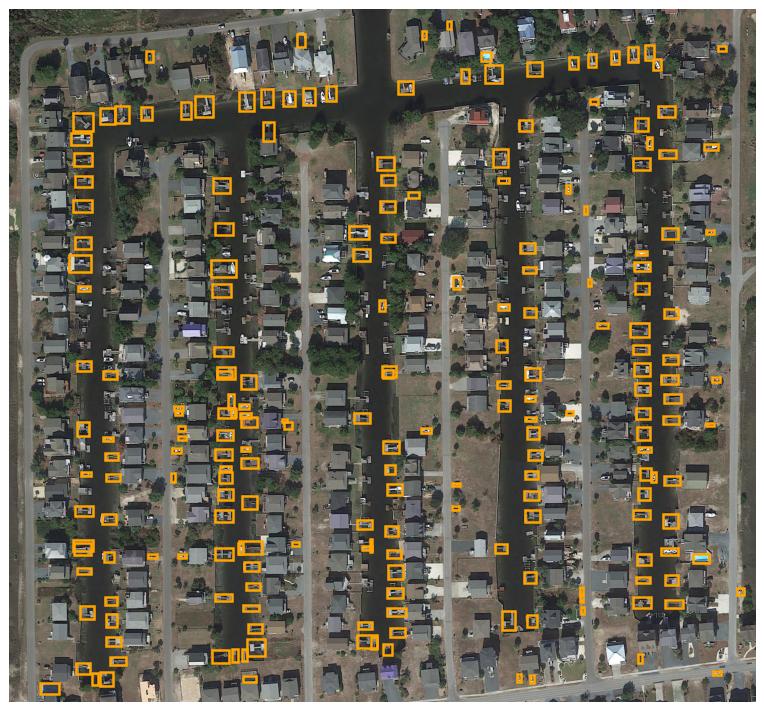} &
\includegraphics[width=0.25\linewidth, height=3.2cm]{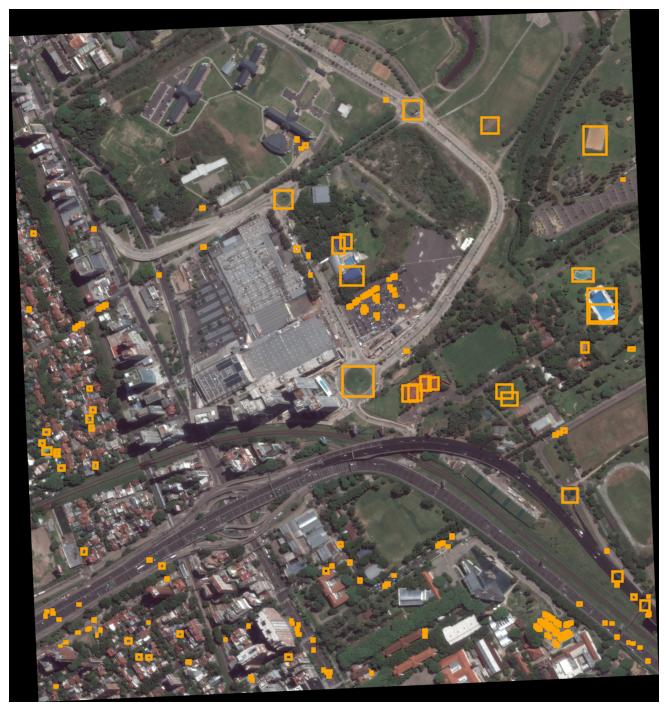} &
\includegraphics[width=0.25\linewidth, height=3.2cm]{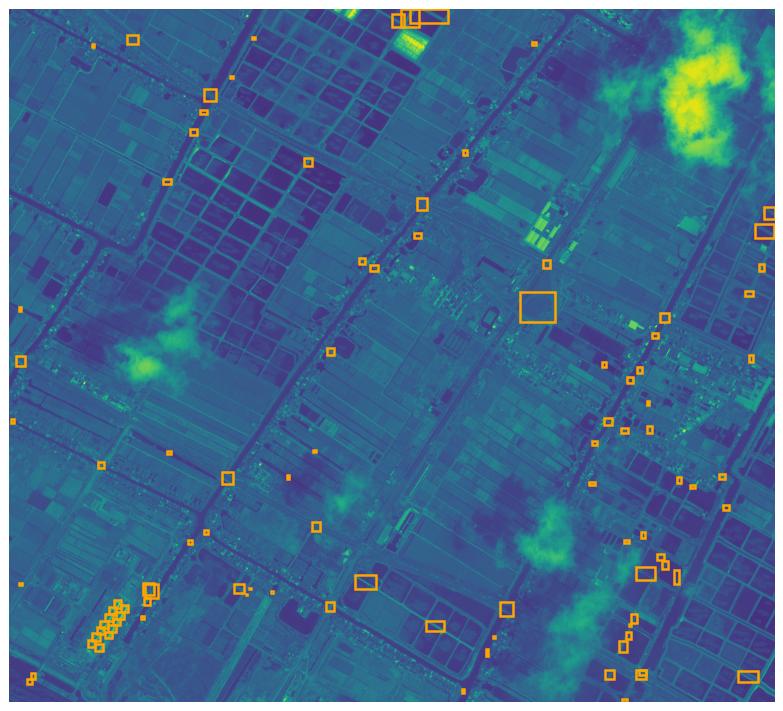} & 
\includegraphics[width=0.25\linewidth, height=3.2cm]{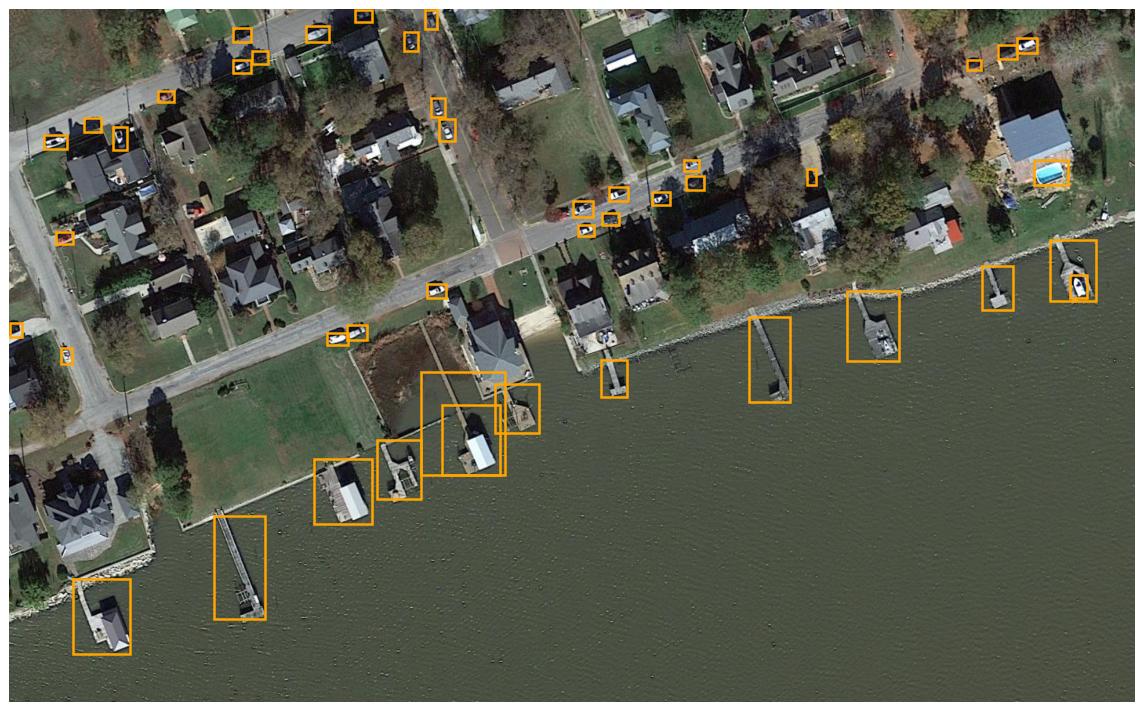}\\
\rotatebox{90}{\quad Sup. VisDrone} & \includegraphics[width=0.25\linewidth, height=3.0cm]{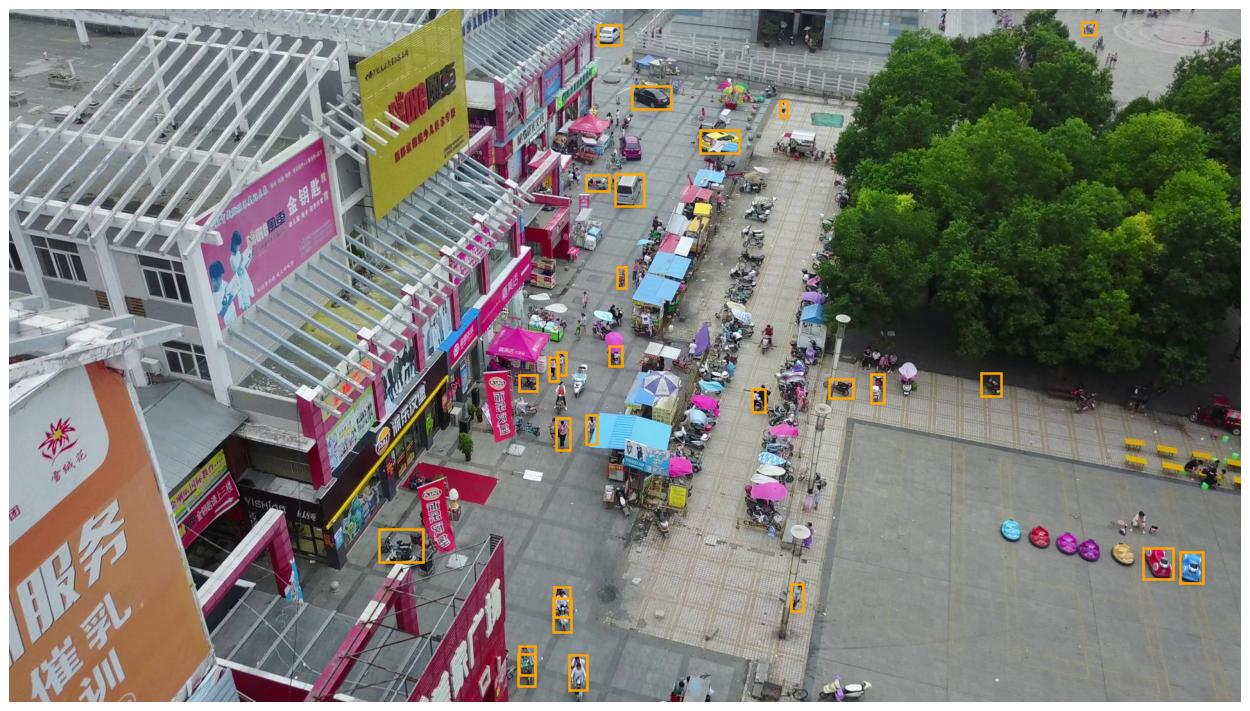} &
\includegraphics[width=0.25\linewidth, height=3.0cm]{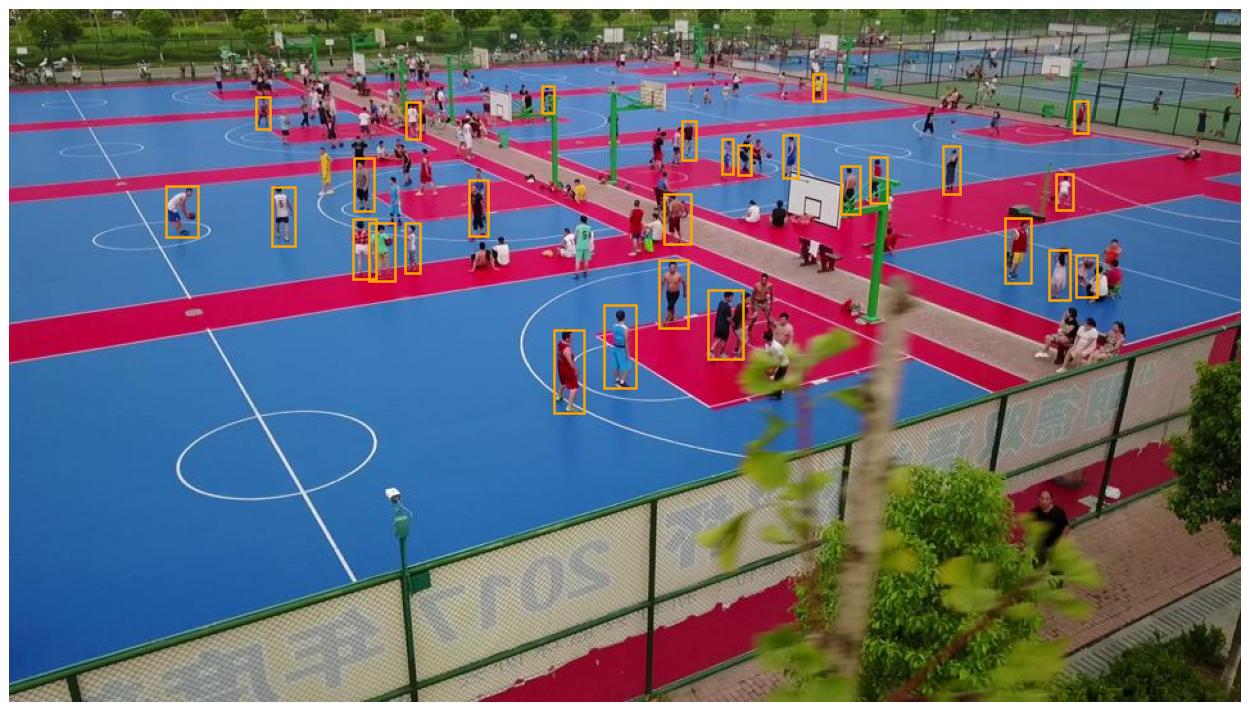} &
\includegraphics[width=0.25\linewidth, height=3.0cm]{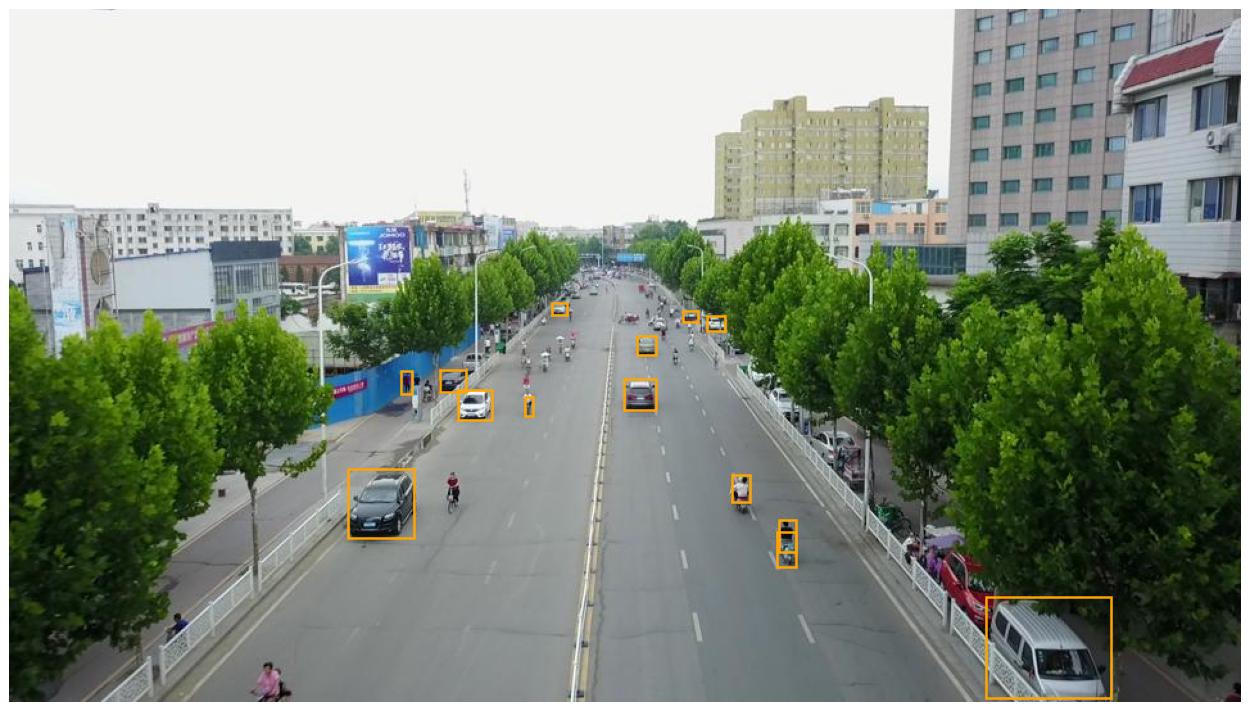} &
\includegraphics[width=0.25\linewidth, height=3.0cm]{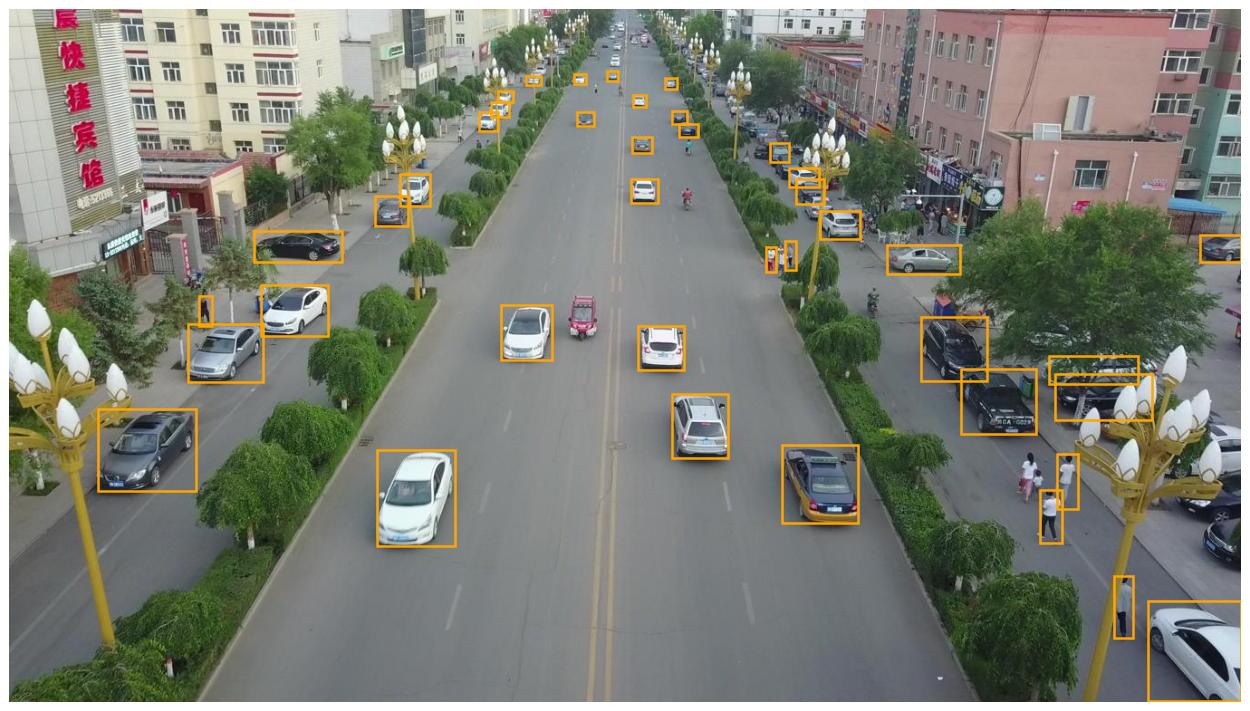}\\
\rotatebox{90}{\quad SSOD+C VisDrone} & \includegraphics[width=0.25\linewidth, height=3.0cm]{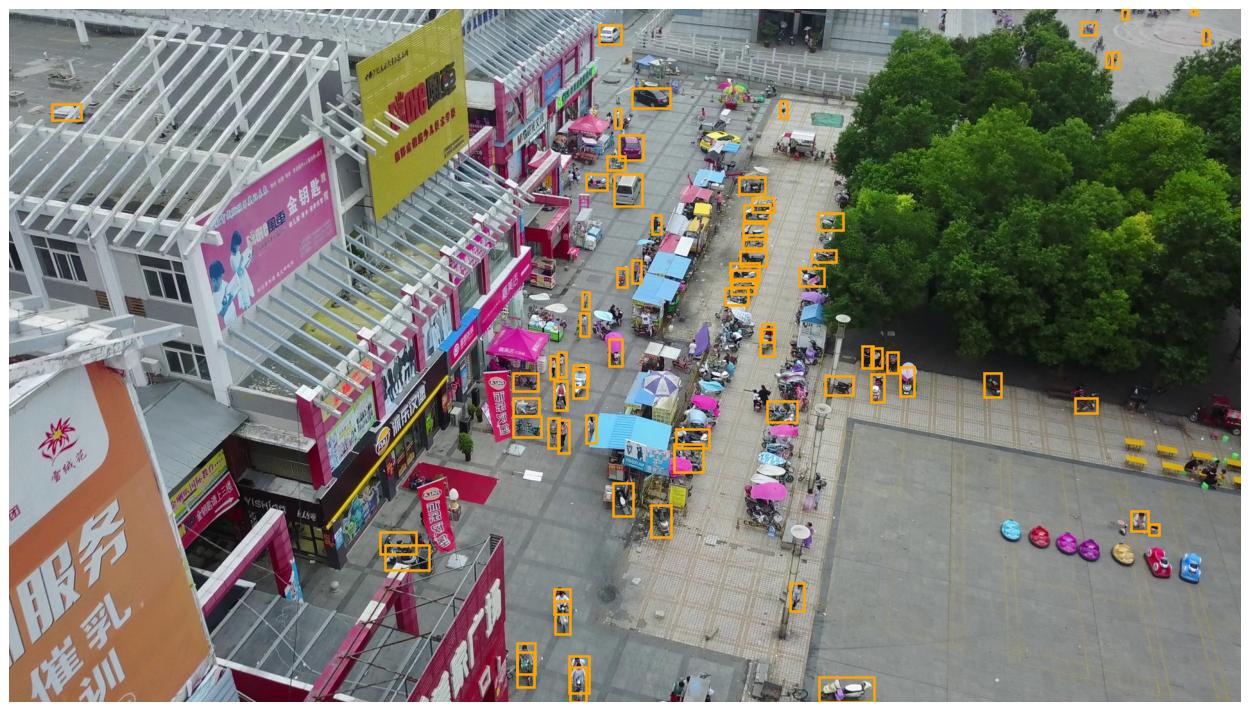} &
\includegraphics[width=0.25\linewidth, height=3.0cm]{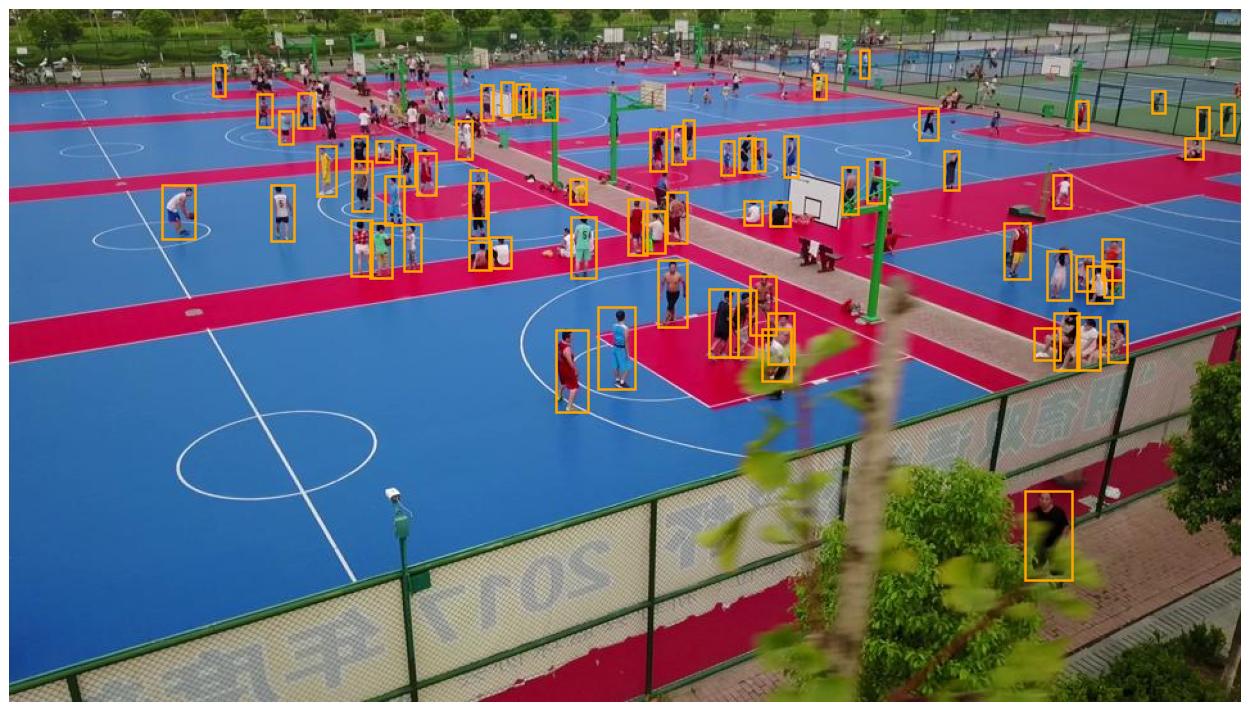} &
\includegraphics[width=0.25\linewidth, height=3.0cm]{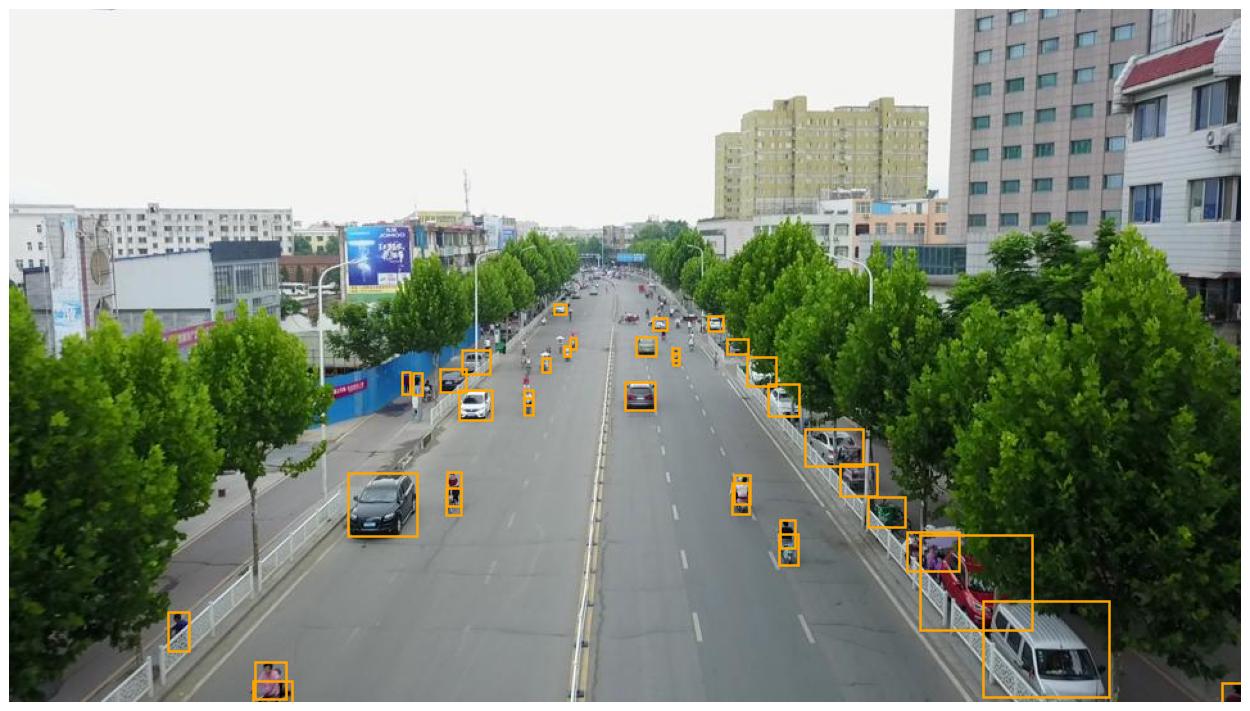} &
\includegraphics[width=0.25\linewidth, height=3.0cm]{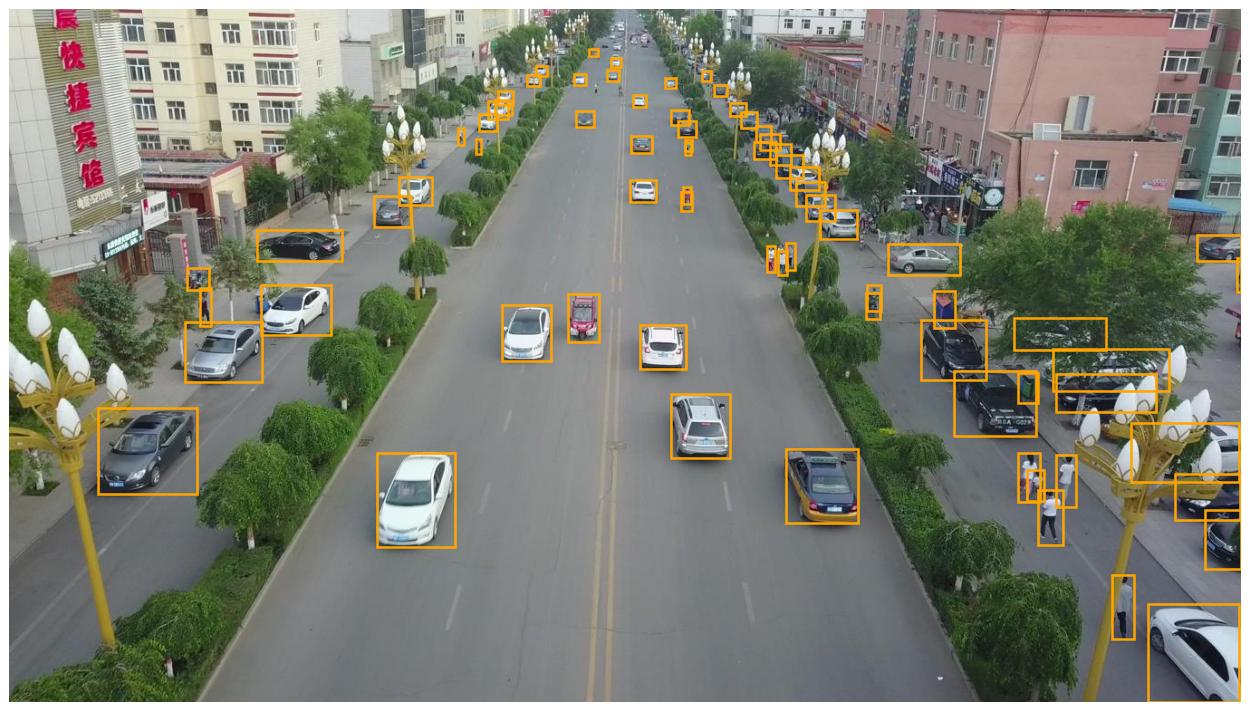}\\
 \end{tabular}
 \caption{Qualitative comparison of detection results between supervised baseline and semi-supervised detector trained with density crops. More objects are detected with our semi-supervised zoom-in detector, especially the small ones. }
 \label{fig:det_comparison}
\end{figure*}

We also produced a qualitative comparison of the detection results from our semi-supervised model with that of its supervised baseline. Figure \ref{fig:det_comparison} shows the comparison on the DOTA (top two rows) and VisDrone (bottom two rows) datasets. The supervised baseline is shown at the top and the semi-supervised results at the bottom among each pair of rows. We can see that many tiny objects are getting detected with our density-guided semi-supervised detector. In the case of VisDrone datasets, the baseline detector is missing most of the small objects at the farther end of the camera, whereas our method with zoom-in capability is discovering them. In the DOTA dataset, the missing happens at a much higher rate as the images are very high in pixel resolution. Especially objects like small cars are mostly missed by the baseline detector on the DOTA dataset. But our method shows impressive results in detecting them.

\subsection{Comparison with Other Semi-supervised Detectors}
As other density-based approaches for small object detection use an external module (and multi-stage training) for crop extraction, we cannot adapt them to the semi-supervised settings with mean-teacher. So, we choose the recently proposed scale-aware detection QueryDet \cite{querydet-Yang-2022} as it also accelerates small object detection with a detector itself. In particular, they proposed sparse querying on the high-resolution feature maps to improve small object detection. This is implemented on the feature pyramids within a detector, so we can wrap the mean-teacher training on top of this method. We used the VisDrone dataset with 10\% labels in this study. The result is shown in table \ref{table:other_ssod_comp}. Our method has an AP of more than 7\% compared to the QueryDet semi-supervised detector. The $\textrm{AP}_{s}$ is improved by 7\% whereas $\textrm{AP}_{m}$, $\textrm{AP}_{l}$ has an improvement of more than 10\%. While the semi-supervised QDet has an improvement of 3\% over its supervised baseline, our method has an improvement of 8\% over the supervised baseline. Note that the supervised baselines are different here because QueryDet proposed a method specific to the RetinaNet \cite{retinanet-Lin-2017} detector. This study also establishes the superiority of density-based detection over scale-aware training as well. 

\begin{table}
    \caption{Performance comparison with QueryDet \cite{querydet-Yang-2022} method for small object detection in the semi-supervised settings using 10\% labeled images on the VisDrone dataset.}
    \centering
    \resizebox{.48\textwidth}{!}{% <------ Don't forget this %
      \begin{tabular}{r||rrr|rrr}
    \hline
    \textbf{Settings} & \textbf{AP} & \textbf{$\textrm{AP}_{50}$} & \textbf{$\textrm{AP}_{75}$} & \textbf{$\textrm{AP}_{s}$}  & \textbf{$\textrm{AP}_{m}$} & \textbf{$\textrm{AP}_{l}$} \\ 
    \hline
    Sup. QDet\cite{querydet-Yang-2022} & 16.58 & 31.13 & 15.45 & 10.89 & 23.93 & 23.46 \\
    Sup. Ours & 19.26 & 37.73 & 17.48 & 12.94	& 26.85	& 26.65 \\
    \hline
    QDet\cite{querydet-Yang-2022} SSOD & 19.56 & 35.78 & 18.67 & 13.90 & 26.17 & 30.53 \\
    \rowcolor{mygray} Ours & \textbf{27.46} & \textbf{48.95} & \textbf{26.92} & \textbf{19.88} & \textbf{37.73} & \textbf{36.31} \\
    \hline
    \end{tabular}% <------ Don't forget this %
    }
    \label{table:other_ssod_comp}
\end{table}

\subsection{Ablation Studies}
The improvement in performance when progressively adding density crops and semi-supervised detection is verified with the results in tables \ref{table:diff_sup_data_dota} and \ref{table:diff_sup_data_visdrone}. The results in these tables ablate the components of the proposed method extensively. Furthermore, the change in performance on objects of different sizes with these components is illustrated in figures \ref{fig:vis_sml} and \ref{fig:dota_sml}. The observations conclude that using density crops with the semi-supervised mean-teacher detector significantly improves the results over the basic supervised detector. As the impact of the components of the proposed method is verified with these experiments, we devote the ablation studies to consider other design choices and fine-grained analysis of the method's performance. We used the VisDrone dataset in this study.

\subsection{Inference}
The inference with density crops can be performed in two ways; taking the crop prediction directly from the model or running the cluster labeling algorithm with output detections. While the crop predictions are fast for inference, we observed that running the cluster labeling algorithm on the detection output is slightly more accurate. So one can choose the inference procedure among the two based on the speed vs accuracy trade-off of the downstream application. In the results reported so far, we used crop predictions directly from the model. To compare the performance of both we performed inference in two ways and reported the performance in table \ref{table:inf_comparison}. The VisDrone dataset with 10\% labels is used in this study. We can observe that while the improvement is small in AP, $\textrm{AP}_{50}$ has a gain of more than 1\%. We can also see that crop-labeled inference is improving the AP of small objects significantly, but at the same time, the AP of medium and large objects is declining. As the dataset is dominated by small objects, we still observe an overall improvement in performance. We also reported the detection speed in Frames Per Second (FPS). The FPS is only reduced by 5 frames when the expensive crop-labeled inference is used.

\begin{table}
    \caption{Results comparison of inference with predicted crops vs crops obtained by algorithm \ref{alg:crop_discovery} on detection boxes. The Visdrone dataset with 10\% labels is used in the study.}
    \centering
    \resizebox{.48\textwidth}{!}{% <------ Don't forget this %
      \begin{tabular}{r||rrr|rrr|r}
    \hline  
    \textbf{Settings} & \textbf{AP} & \textbf{$\textrm{AP}_{50}$} & \textbf{$\textrm{AP}_{75}$} & \textbf{$\textrm{AP}_{s}$}  & \textbf{$\textrm{AP}_{m}$} & \textbf{$\textrm{AP}_{l}$} & \textbf{FPS} \\  
    \hline
    \shortstack{Inference with \\ Predicted crops} & 27.46 & 48.95 & 26.92 & 19.88 & 37.73 & 36.31 & 12.45\\
    \shortstack{Inference with \\ labeled crops} & 27.78 & 50.02 & 27.12 & 20.99 & 36.32 & 35.84 & 7.17\\
    \hline
    \end{tabular}% <------ Don't forget this %
    }
    \label{table:inf_comparison}
\end{table}

\subsection{Comparison with the Supervised Upper-bound}
In table \ref{table:sup_upperbound_ssod_comp}, we compare the results of our semi-supervised model with the fully supervised upper bound where 100\% images are labeled. The setting used here is images labeled 10\%. The lower bound of the performance when only the available 10\% labeled data is also provided. It can be observed that our method with 10\% labeled data is approximately 6\% points close to the upper bound, both in the AP and $\textrm{AP}_{s}$. $\textrm{AP}_{m}$ and $\textrm{AP}_{l}$ also show a similar trend. Therefore, it can be concluded that, by effectively leveraging unlabeled data, our method is able to achieve a performance close to the fully supervised upper bound, while using minimal labeled data points.

\begin{table}
    \caption{Performance comparison with the fully supervised upper-bound on the VisDrone dataset with 10\% labeled images.}
    \centering
    \resizebox{.48\textwidth}{!}{% <------ Don't forget this %
      \begin{tabular}{r||rrr|rrr}
    \hline
    \textbf{Settings} & \textbf{AP} & \textbf{$\textrm{AP}_{50}$} & \textbf{$\textrm{AP}_{75}$} & \textbf{$\textrm{AP}_{s}$}  & \textbf{$\textrm{AP}_{m}$} & \textbf{$\textrm{AP}_{l}$} \\ 
    \hline
    \emph{Lower bound} & & & & & & \\
    10\% Labeled & 19.26 & 37.73 & 17.48 & 12.94 & 26.85 & 26.65 \\
    \hline
    \emph{Semi-supervised (Ours)} & & & & & & \\
    10\% Lab. + 90\% Unlab. & 27.46 & 48.95 & 26.92 & 19.88 & 37.73 & 36.31 \\
    \hline
    \emph{Upper bound} & & & & & & \\
    100\% Labeled & 33.22 & 58.30 & 33.16 & 26.06 & 42.58 & 43.36 \\
    \hline
    \end{tabular}% <------ Don't forget this %
    }
    \label{table:sup_upperbound_ssod_comp}
\end{table}
%%%%%%%%%

\subsection{computational cost}
Using unlabeled data for mean-teacher training comes with additional training costs. Exponentially averaged teacher weights must be learned with a small $\alpha$ value to have stable distillation. We used a 0.9996 following the standard practices \cite{unbiased-teacher-Liu-2021, humble-teacher-Yang-2021}. This results in many iterations for the mean-teacher training. In table \ref{table:cost_comparison}, we compared the training iterations and time for different settings. Inference time per image is also provided. Finding crops on unlabeled images is performed only after the pseudo labels on unlabeled images are converged. The augmentation then adds an additional set of crops to the training process. That is why the SS+C (L+U) setting is taking longer iterations. For inference, the difference when using crops is due to the second detection performed on the crops. Even though there is an effective increase in training and inference time, the improvement in detection performance is significant.

\begin{table}
    \caption{Comparison of the training and test time for fully supervised and semi-supervised methods with and without density crops. All settings are evaluated using one A100 GPU with the Visdrone dataset having 10\% labels.}
    \centering
    \resizebox{.48\textwidth}{!}{% <------ Don't forget this %
      \begin{tabular}{r||rrrrr}
    \hline  
    \textbf{Settings} & \textbf{FS} & \textbf{FS+C} & \textbf{SS} & \textbf{SS+C (L)}  & \textbf{SS+C (L+U)} \\ 
    \hline
    \shortstack{Train iters} & 5k & 15k & 65k & 75k & 180k\\
    \shortstack{Train time in HH:MM} & 1:03 & 2:28 & 15:36 & 15:19 & 33:35\\
    \shortstack{Test time in s/image} & 0.0348 & 0.0661 & 0.0348 & 0.0661 & 0.0661 \\
    \hline
    \end{tabular}% <------ Don't forget this %
    }
    \label{table:cost_comparison}
\end{table}

\subsection{Analysis of the Type of Errors}
\begin{figure*}
\centering
\begin{tabular}{@{\hspace{0.2cm}} c @{\hspace{0.3cm}} c @{\hspace{0.3cm}} c @{\hspace{0.3cm}} c}
\includegraphics[width=0.23\linewidth, height=6cm]{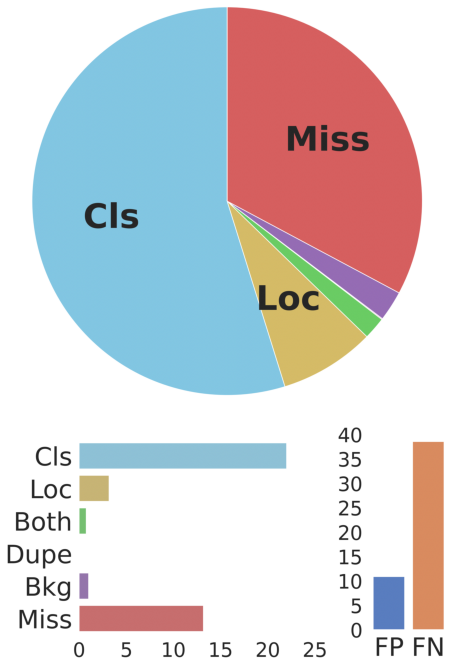} &
\includegraphics[width=0.23\linewidth, height=6cm]{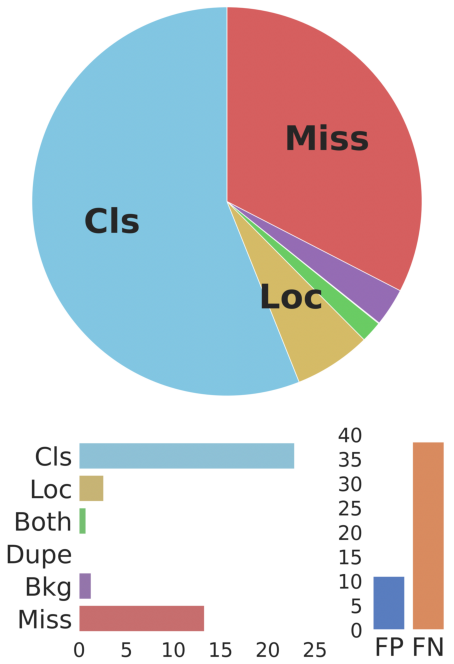} &
\includegraphics[width=0.23\linewidth, height=6cm]{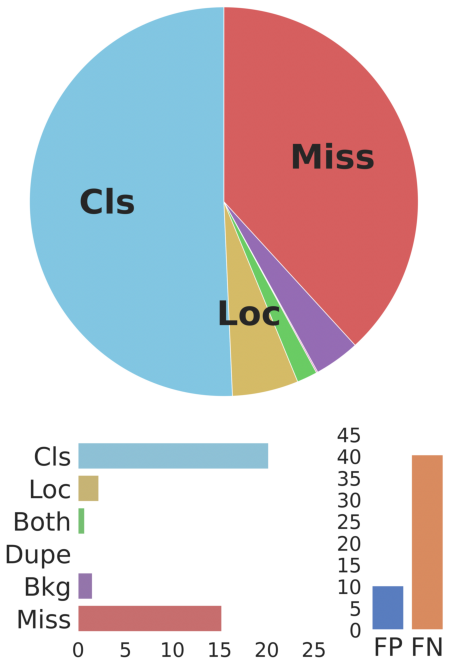} &
\includegraphics[width=0.23\linewidth, height=6cm]{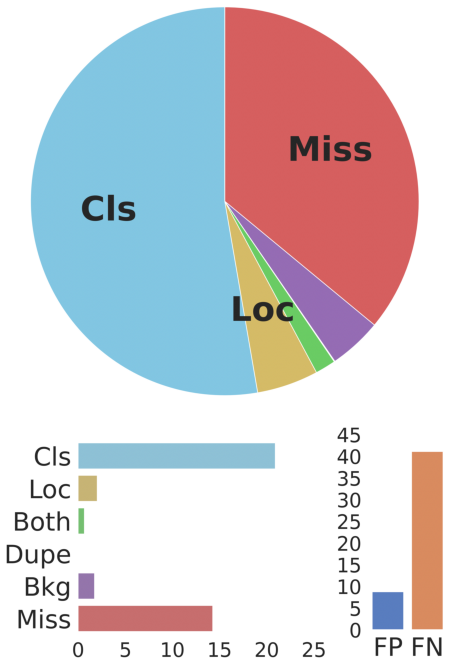}\\
 (a) Supervised & (b) Supervised + Crop  & (c) SSOD & (d) SSOD + Crop \\
 \end{tabular}
 \caption{TIDE\cite{tide-Bolya-eccv2020} evaluation of detection results of the detectors trained with (a) supervised, (b) supervised with density crops, (c) vanilla semi-supervised and (d) semi-supervised with density crops modes. Error types are: \textbf{Cls}: localized correctly but classified incorrectly, \textbf{Loc}: classified correctly but localized incorrectly, \textbf{Both}: both cls and loc error, \textbf{Dupe}: duplicate detection error, \textbf{Bkg}: detected background as foreground, \textbf{Miss}: missed ground truth error.}
 \label{fig:tide_comparison}
\end{figure*}

To understand how the addition of semi-supervised learning and density crops affects the detector's abilities, we profiled different error types based on the TIDE \cite{tide-Bolya-eccv2020} evaluation protocol. Figure \ref{fig:tide_comparison} shows the comparison results. With the addition of density crops on a supervised detector, we observe the localization error reduces. Other types of errors remain mostly the same. With semi-supervised training using the vanilla mean-teacher method, the classification error reduces. Using density crops with semi-supervised learning is reducing the localization error  similar to the fully-supervised case and other errors remain the same mostly. Compared to fully supervised detectors, semi-supervised detectors reduce classification error, but they tend to miss objects too. This is probably due to the imbalance in object classes of this dataset such that dominant classes get more pseudo-labels on unlabeled images. This can result in rare class objects being missed on the unlabeled images.

% !TEX root=main.tex

\section{Conclusion}
\label{sec:conclusion}
We proposed an efficient adaptation of the mean-teacher semi-supervised method to high-resolution aerial images for the detection of small objects. This is achieved by identifying the clusters of small objects from labeled and unlabeled images and processing them in higher resolution. For the labeled images, the original ground-truth is used for crop identification, whereas on unlabeled images the pseudo ground-truth labels from the mean-teacher detector are used. As crop identification is happening within the detector, it is now possible to wrap the mean-teacher training on top of it. The clusters identified are cropped and used to augment the training set. The training with augmented crops is producing more pseudo-labels than the vanilla mean-teacher. This translates to improved detection performance. The inference is performed on the original image and crops of clusters obtained on it to boost the small object detection. Empirical studies on the popular benchmark datasets  reveal the superiority of our method over supervised training and vanilla mean-teacher training. We also find more boost in performance for density-based approaches than the scale-aware training with the mean-teacher method for small object detection.

{\small
\bibliographystyle{ieee_fullname}
\bibliography{bare_jrnl_TGRS}
}

\vfill

\end{document}